\documentclass[final]{cvpr}

\usepackage{times}
\usepackage{epsfig}
\usepackage{graphicx}
\usepackage{amsmath}
\usepackage{amssymb}

\usepackage[pagebackref=true,breaklinks=true,colorlinks,bookmarks=false]{hyperref}

\def\cvprPaperID{5325} % *** Enter the CVPR Paper ID here
\def\confYear{CVPR 2021}

\begin{document}

%%%%%%%%% TITLE
\title{Robustness of Facial Recognition to GAN-based Face-morphing Attacks}

%\author{\parbox{16cm}{\centering
%    {\large Richard T. Marriott$^{1,2}$, Sami Romdhani$^2$, St\'ephane Gentric$^2$ and Liming Chen$^1$}\\
%    {\normalsize
%    $^1$ Ecole Centrale de Lyon, Ecully, France\\
%    $^2$ IDEMIA, Courbevoie, France}}
%    %\thanks{This work was not supported by any organization}% <-this % stops a space
%}

\author{Richard T. Marriott$^{1,2}$\qquad Sami Romdhani$^1$\qquad St\'ephane Gentric$^1$ \qquad Liming Chen$^2$\\
{\normalsize
$^1$ IDEMIA, France\qquad $^2$ Ecole Centrale de Lyon, France}\\
{\tt\small richard.marriott@idemia.com}
}

\maketitle
\thispagestyle{empty}

%%%%%%%%% ABSTRACT
\begin{abstract}

Face-morphing attacks have been a cause for concern for a number of years. Striving to remain one step ahead of attackers, researchers have proposed many methods of both creating and detecting morphed images. These detection methods, however, have generally proven to be inadequate. In this work we identify two new, GAN-based methods that an attacker may already have in his arsenal. Each method is evaluated against state-of-the-art facial recognition (FR) algorithms and we demonstrate that improvements to the fidelity of FR algorithms do lead to a reduction in the success rate of attacks provided morphed images are considered when setting operational acceptance thresholds.

\end{abstract}

%%%%%%%%% BODY TEXT
\section{Introduction}\label{Sec. Introduction}

The potential threat of morphing attacks to systems secured by facial recognition (FR) was first identified in the 2014 paper ``The Magic Passport'' \cite{DBLP:conf/icb/FerraraFM14}. The paper demonstrated the relative ease with which images can be manipulated to simultaneously resemble multiple identities using commercially available tools, and the vulnerability of FR systems to those images. The extent to which face-morphing as a method of attack has been adopted by criminals is not known since, by definition, successful attacks remain undetected. Nevertheless, a pre-emptive arms race was spawned in the literature, with evermore sophisticated morphing methods being proposed in conjunction with tools for their detection \cite{DBLP:conf/btas/DamerS0K18, DBLP:conf/iwbf/DebiasiSRUB18, DBLP:conf/eusipco/FerraraFM18, DBLP:conf/icisp/ScherhagBGB18, DBLP:conf/eusipco/SeiboldHE18}. Various datasets of morphed examples have been made publicly available \cite{DBLP:conf/eusipco/MahfoudiTRMDP19, DBLP:conf/icb/RaghavendraRVB17} and an ongoing morphing detection benchmark has been included as part of NIST's Face Recognition Vendor Test (FRVT) \cite{ngan2020face}.

In this paper we evaluate the robustness of FR algorithms to two morphing methods that make use of style-based, generative networks; specifically we produce morphed images using StyleGAN pre-trained on the Flickr-Faces-HQ (FFHQ) dataset \cite{DBLP:conf/cvpr/KarrasLA19}. The work of \cite{DBLP:conf/iwbf/VenkateshZRRDB20} evaluates a method similar to the ``midpoint method'' presented here. Whereas \cite{DBLP:conf/iwbf/VenkateshZRRDB20} focusses on assessment of the extent to which FR systems are vulnerable to GAN-based face-morphing attacks in comparison to landmark-based attacks, and also on detection of such attacks, here we focus on the changing response of FR systems to morphed images as fidelity improves. We observe that improvements to FR systems do not necessarily translate to improved robustness to morphing attacks and that morphed images should therefore be taken into account when setting operational acceptance thresholds. We also introduce and evaluate a second, style-based morphing method: the ``dual biometric method''.

\begin{figure}[t]
\begin{center}
   \includegraphics[width=0.33\linewidth]{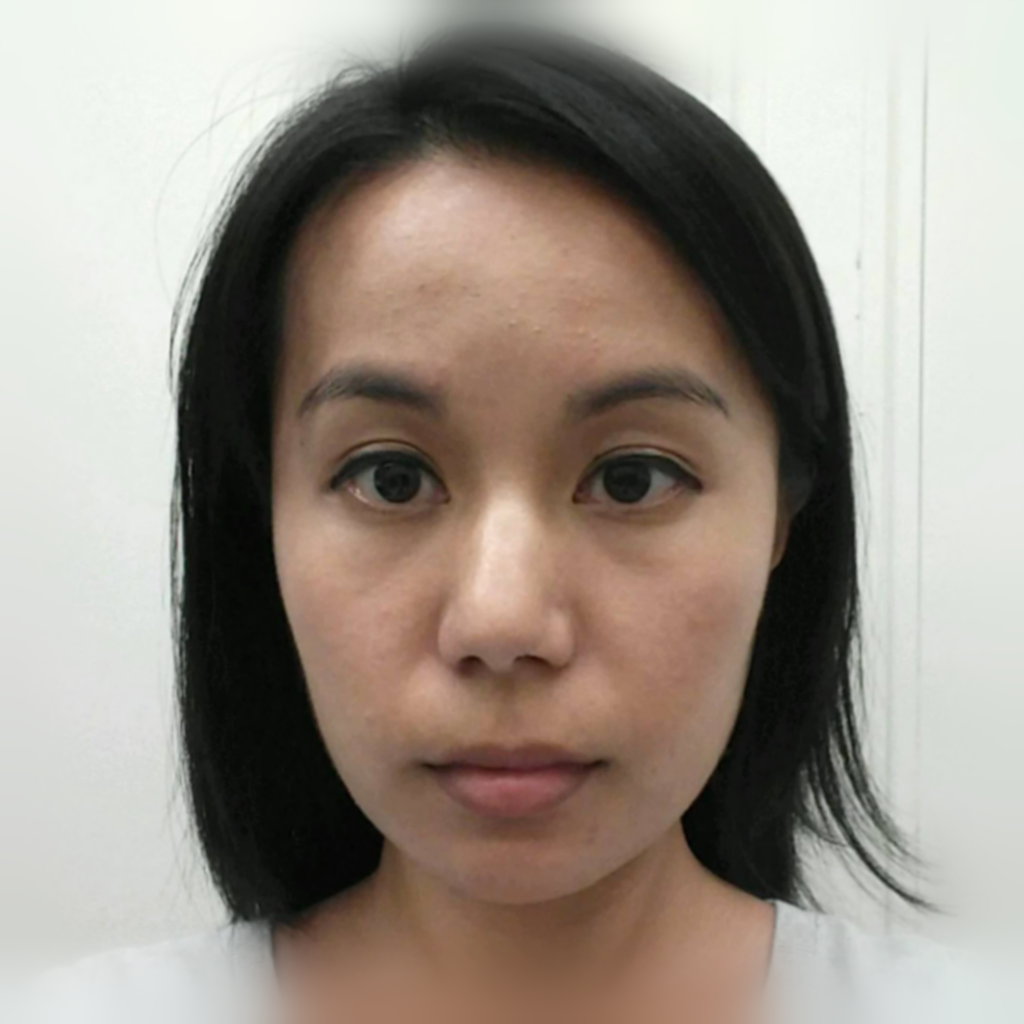}\includegraphics[width=0.33\linewidth]{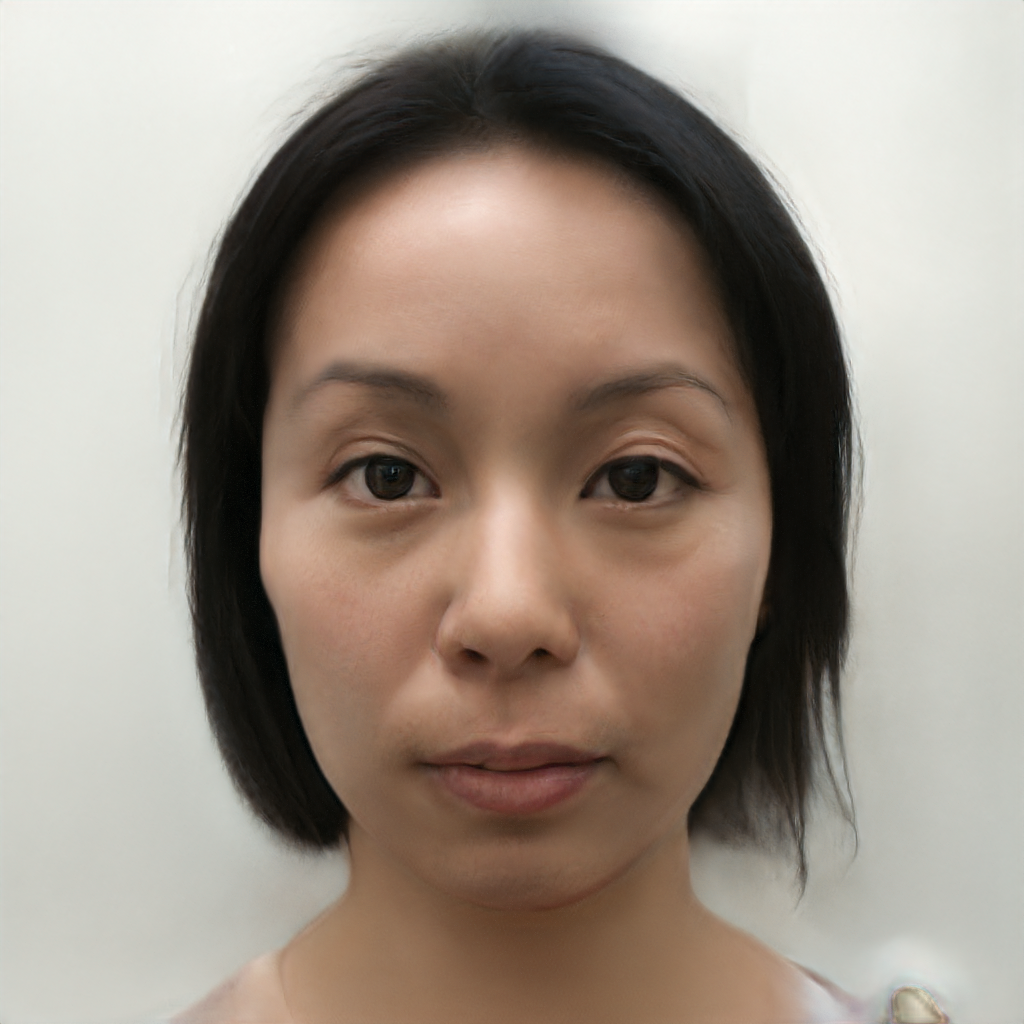}\includegraphics[width=0.33\linewidth]{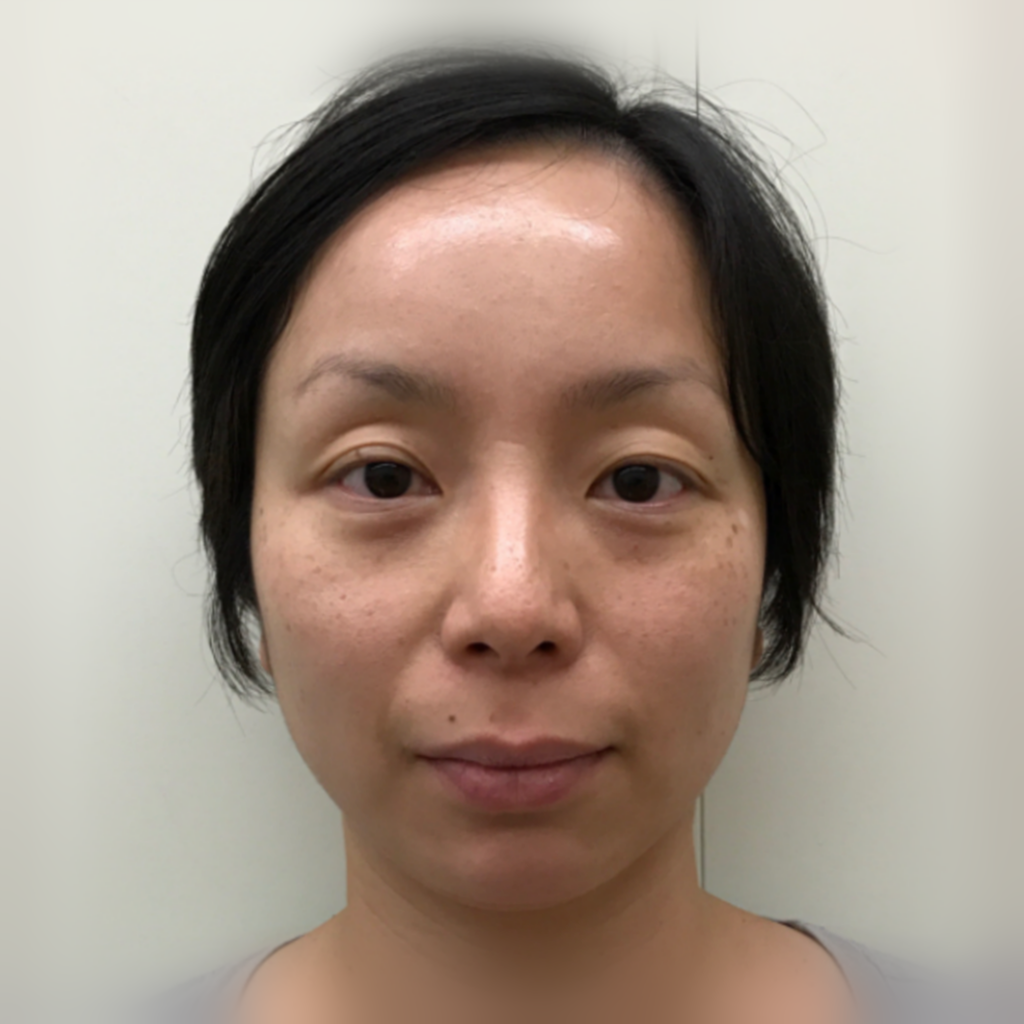}
\end{center}
   \caption{StyleGAN midpoint morph of NIST subjects A and B. Images of subject A (left) and B (right) were taken from \cite{ngan2020face}. The central image is the morph.}
\label{fig:NIST Morph}
\end{figure}

\begin{figure}[t]
\begin{center}
   \includegraphics[width=0.25\linewidth]{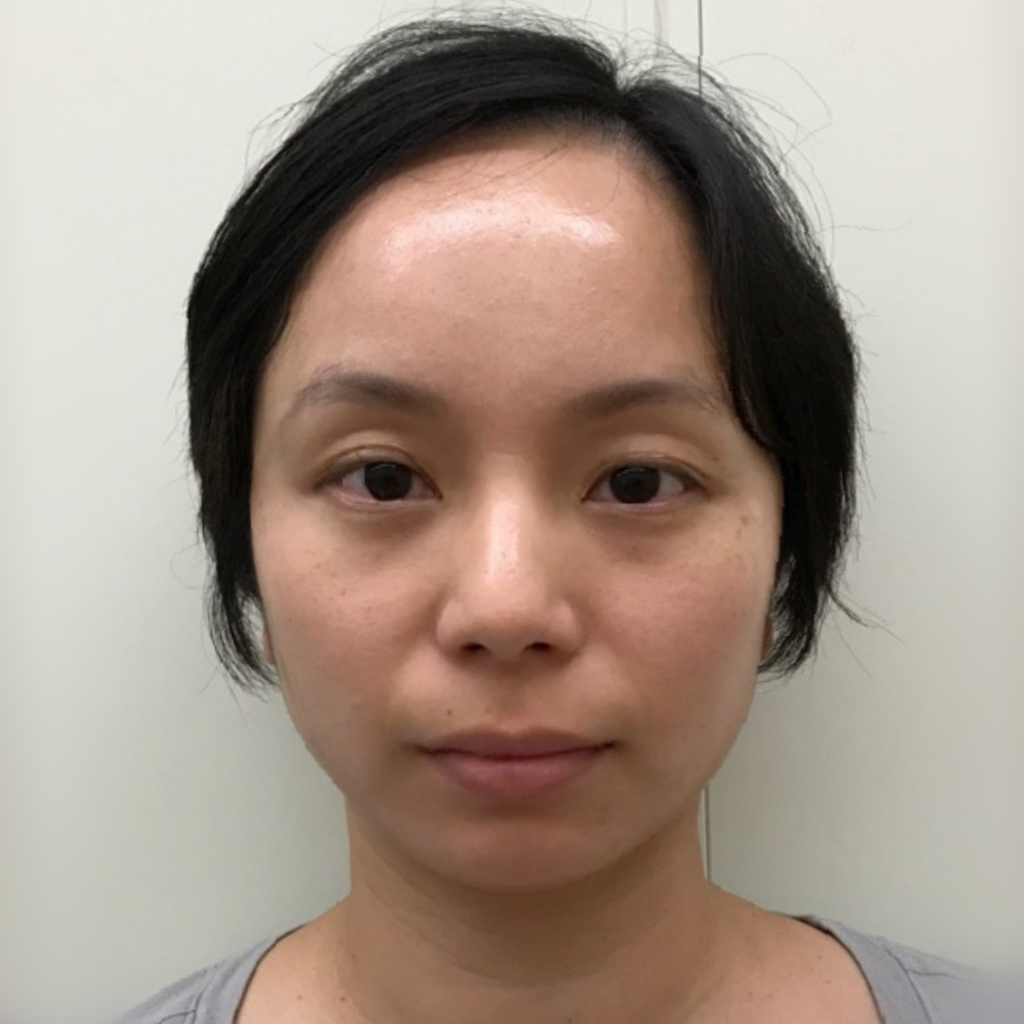}\includegraphics[width=0.25\linewidth]{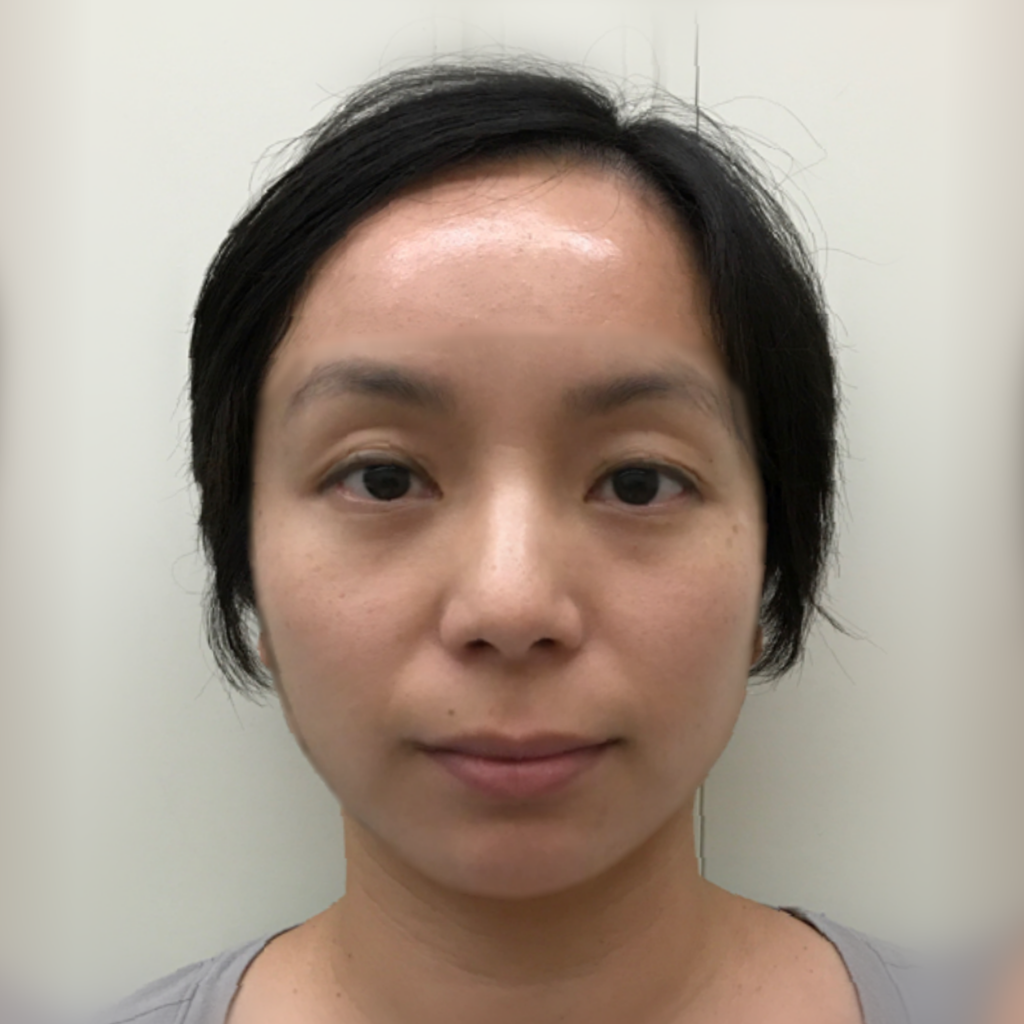}\includegraphics[width=0.25\linewidth]{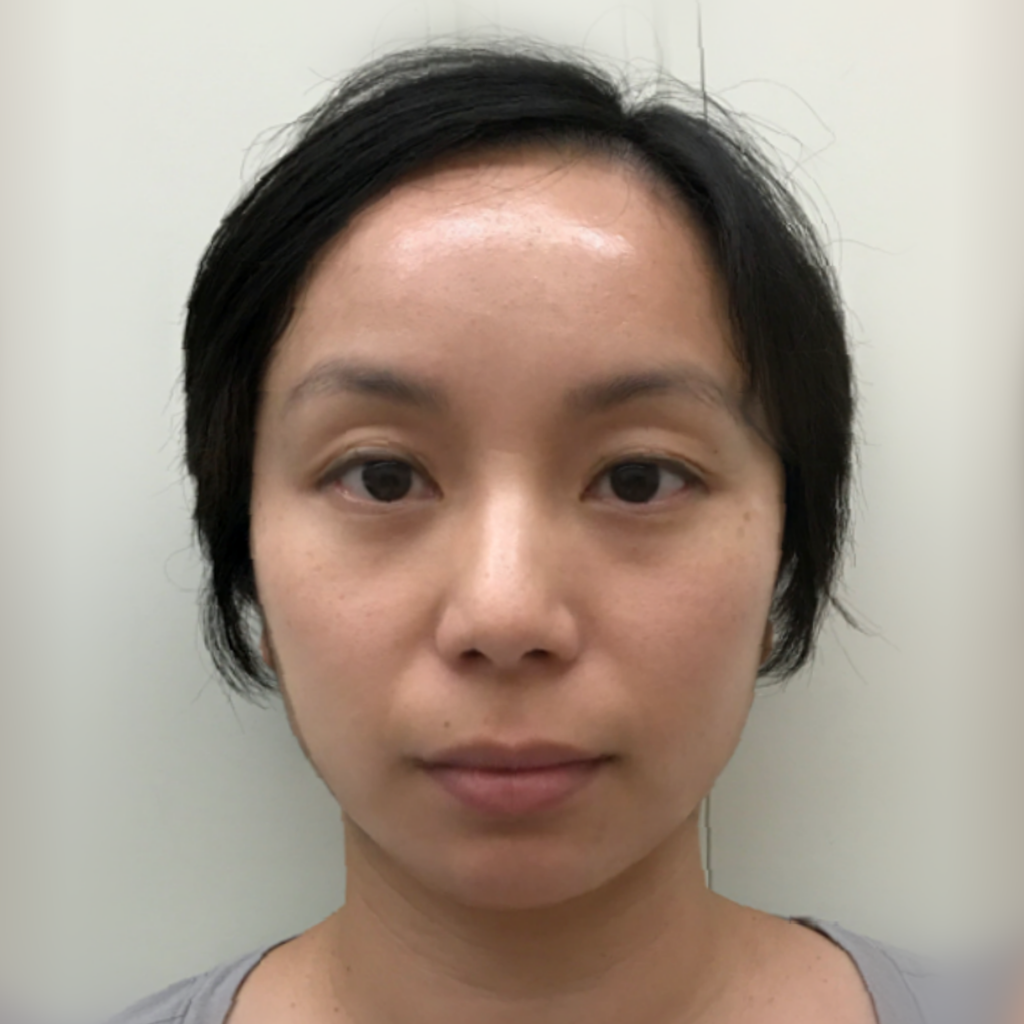}\includegraphics[width=0.25\linewidth]{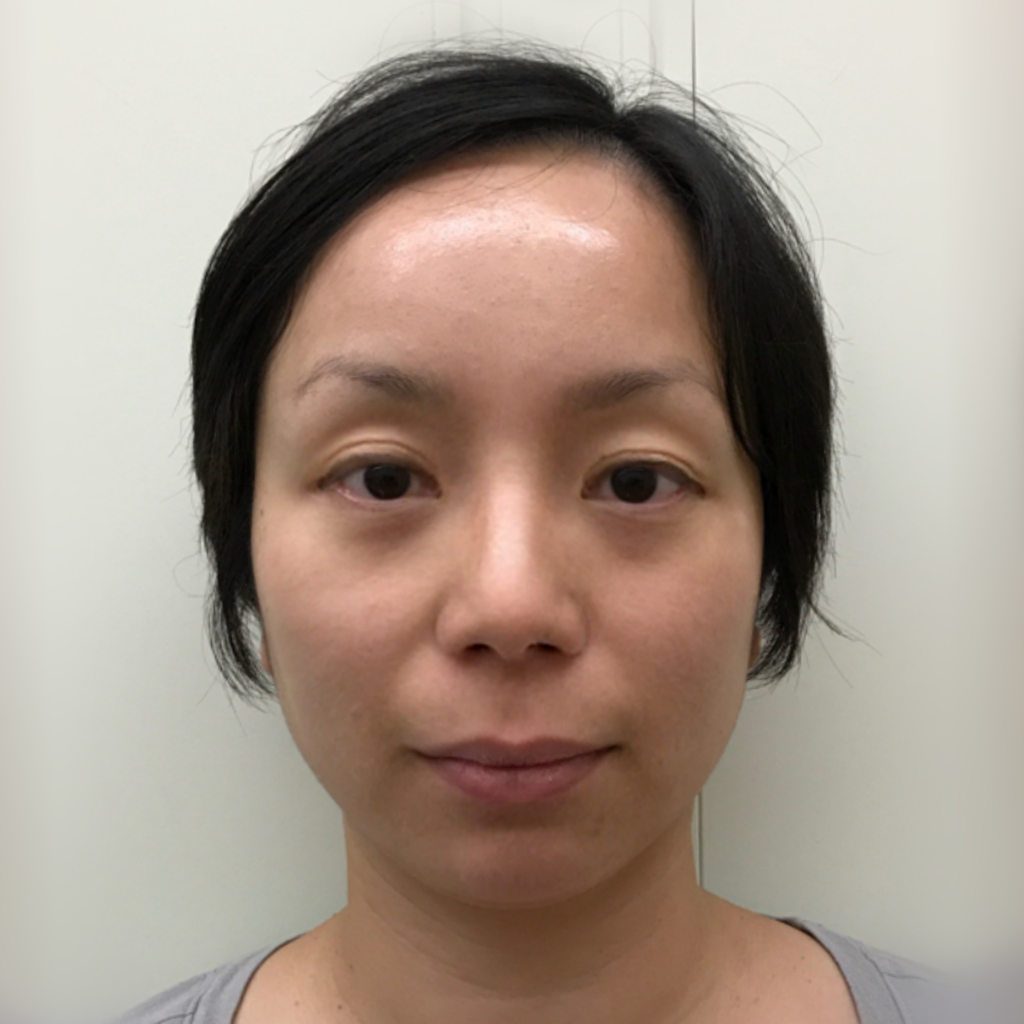}
\end{center}
   \caption{Examples of the output of various alternative, automated morphing methods taken from \cite{ngan2020face}. These correspond to Figure 2 (g), (i), (j) and (l) of the NIST report.}
\label{fig:NIST Auto methods}
\end{figure}

The rest of the paper is organised as follows: in Section \ref{Sec. Morphing attacks} we provide a recap of how morphing attacks proceed and identify potential approaches that might be used to prevent them; in Section \ref{Sec. Related Work} we discuss work in the literature proposing to complement FR systems with algorithms for detection of morphed images, and also work leading to the development of the style-based morphing methods proposed here; in Section \ref{Sec. Face-morphing with StyleGAN} we describe the style-based morphing methods being evaluated, providing results in Section \ref{Sec. Results}; and in Section \ref{Sec. Conclusions} we draw conclusions.

\section{Morphing attacks}\label{Sec. Morphing attacks}

\subsection{How does a face-morphing attack work?}\label{Sec. Intro - How it works}

For the benefit of the uninitiated reader, we recap on how a morphing attack might proceed:

\begin{enumerate}
    \item An accomplice is identified who is willing to share their passport with an imposter. Ideally the accomplice will resemble the imposter.
    \item A morphed image is produced that resembles both the accomplice and the imposter.
    \item The accomplice presents the morphed image at the time of application for the identity document. (The image is found to be plausible and so is accepted.)
    \item The resulting identity document is shared with and used by the imposter.
\end{enumerate}

\subsection{Prevention of morphing attacks}\label{Sec. Intro - Methods of prevention}

There are three broad approaches that might be taken to prevent such an attack:

\begin{enumerate}
    \item \textbf{Trusted capture.} In the prelude to a morphing attack, the accomplice exploits his freedom to provide an image to the issuing authority. Removing this freedom by enforcing live image-capture at the time of application would make attacks significantly more challenging to perpetrate.
    \item \textbf{Morph-detection.} Although currently known morphing methods produce images of high quality, none of them is perfect. Morphed images may contain certain features that betray their dubious provenance. Deploying automated detection of these features, either prior to creation of the identity document or at the time of use, could potentially prevent attacks.
    \item \textbf{Robustness of recognition.} A morphed image contains an identity that is neither that of the accomplice nor of the imposter. A facial recognition system that is effective enough to recognise the identity as such would not be vulnerable to the attack.
\end{enumerate}

In this work we show that the third approach, i.e. working to improve the robustness of facial recognition algorithms to similar but subtly different identities, is an important and effective way of ensuring the integrity of FR-secured systems.

%In this work we look at robustness of recognition and evaluate the effect that recent improvements to the fidelity of FR may have had on the success rate of morphing attacks.

\begin{figure}[t]
\begin{center}
   \includegraphics[width=1.0\linewidth]{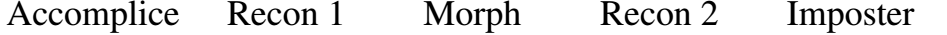}
   \includegraphics[width=0.2\linewidth]{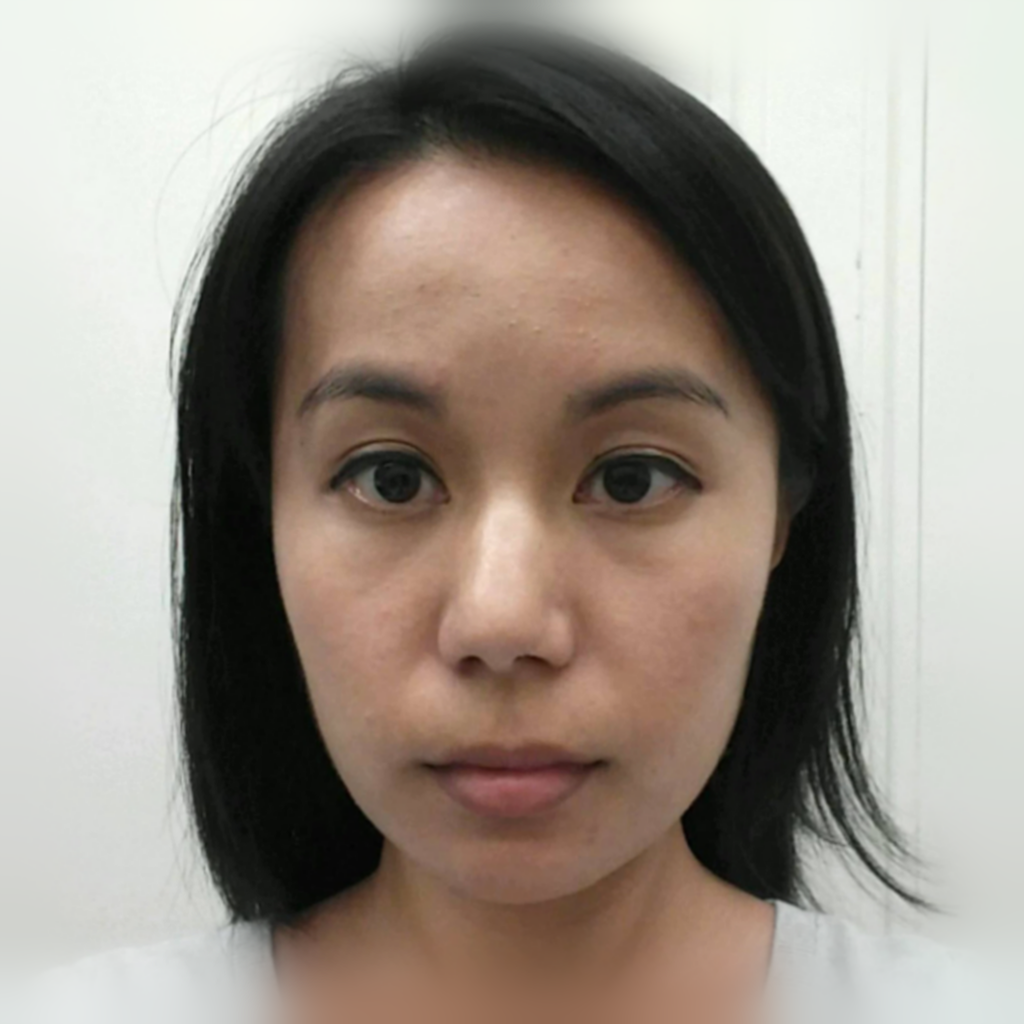}\includegraphics[width=0.2\linewidth]{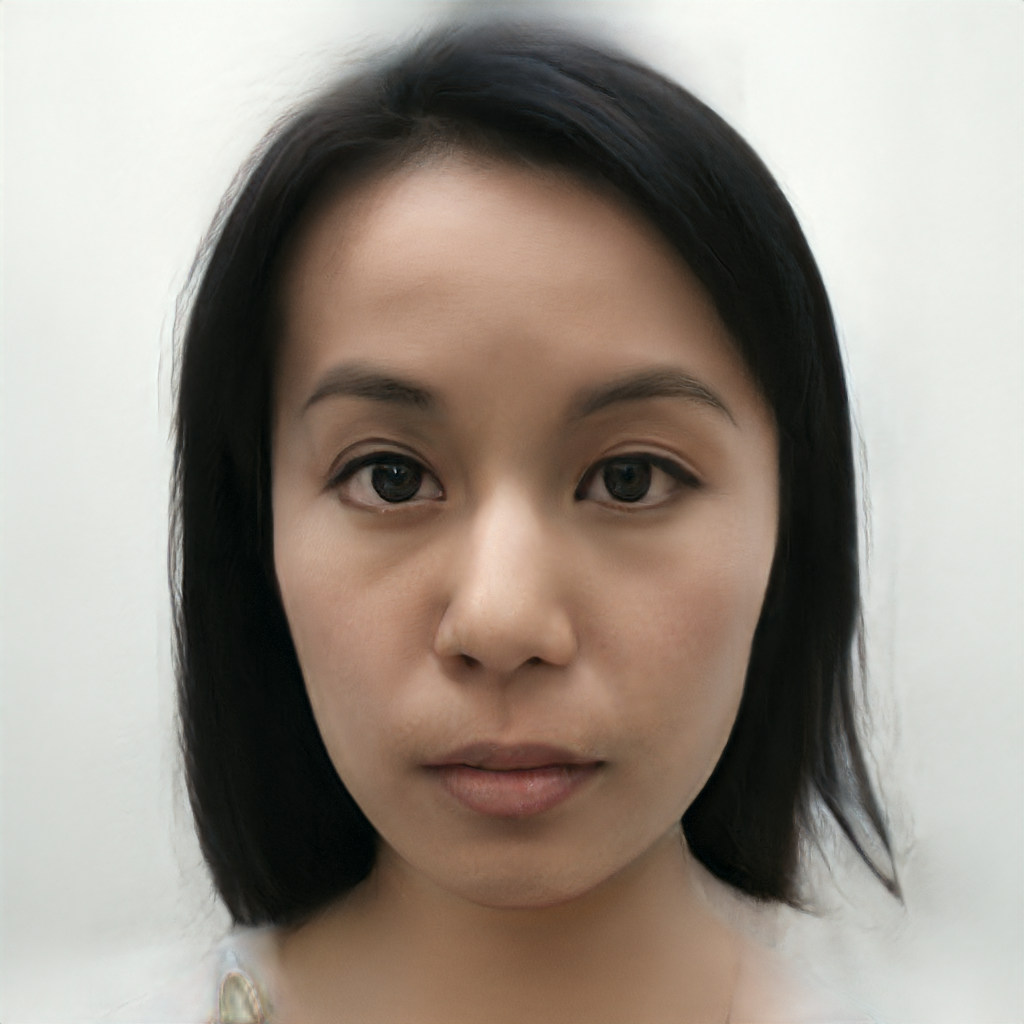}\includegraphics[width=0.2\linewidth]{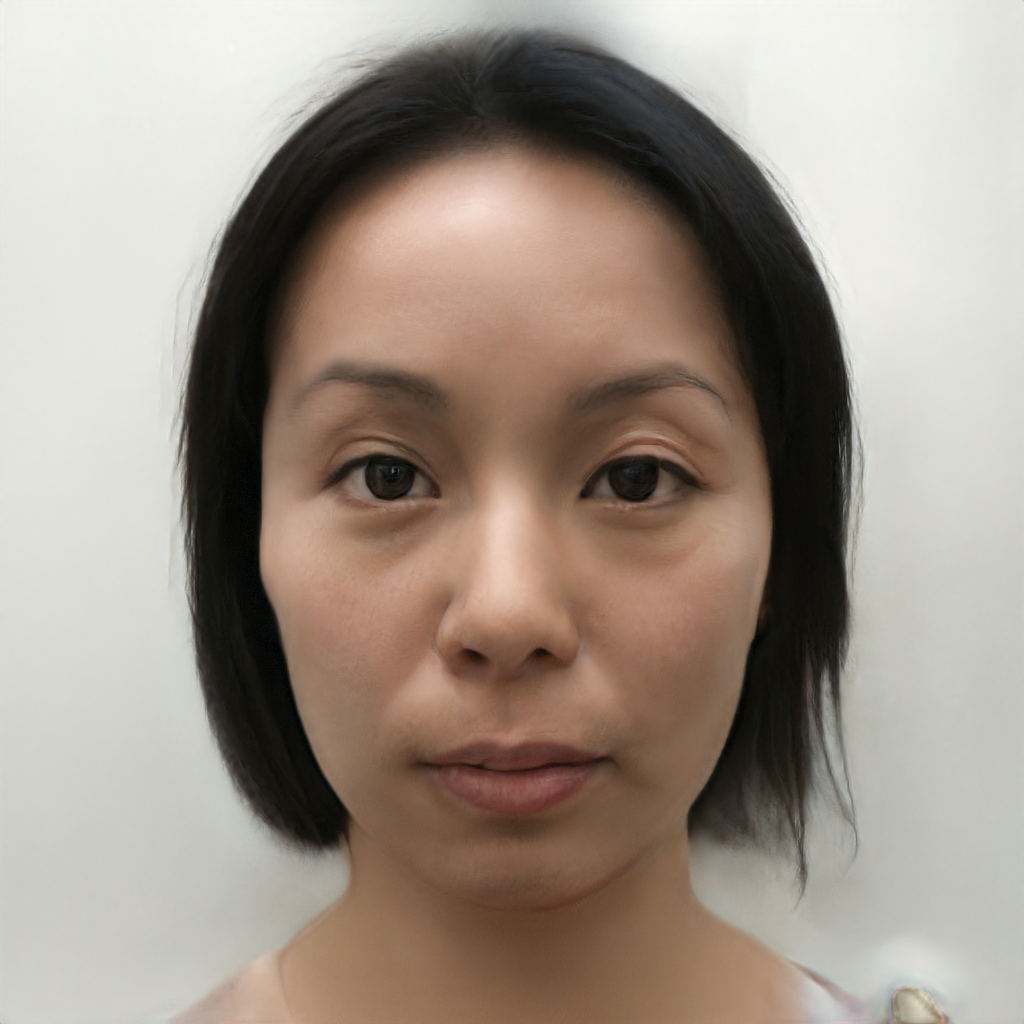}\includegraphics[width=0.2\linewidth]{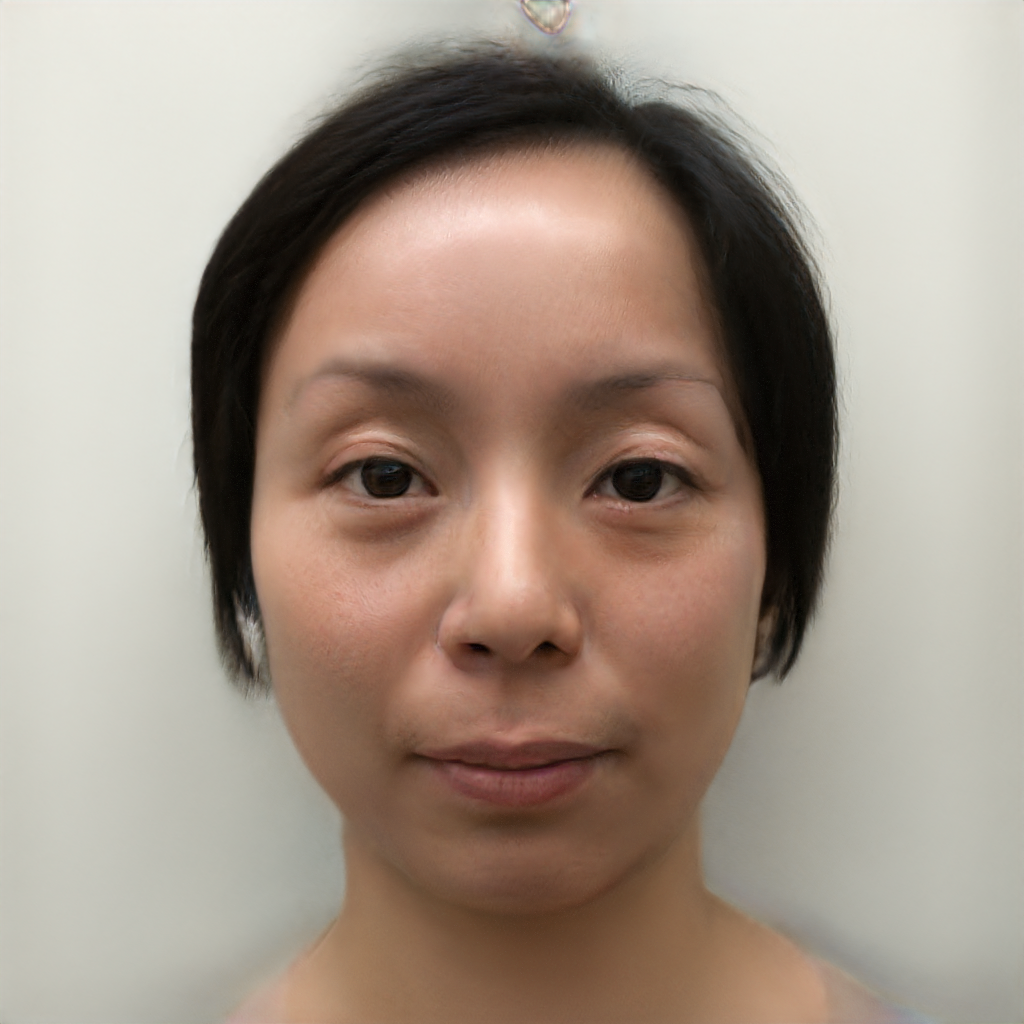}\includegraphics[width=0.2\linewidth]{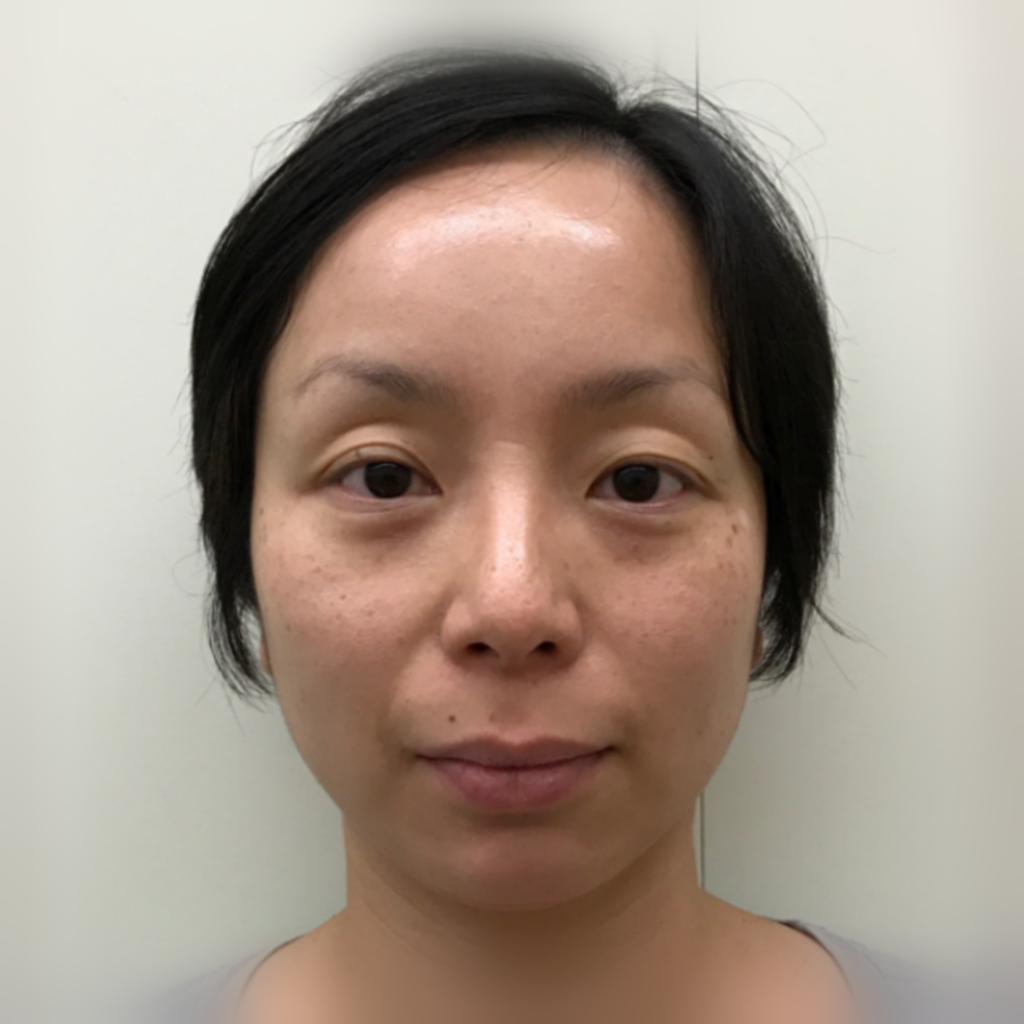}
   \includegraphics[width=0.2\linewidth]{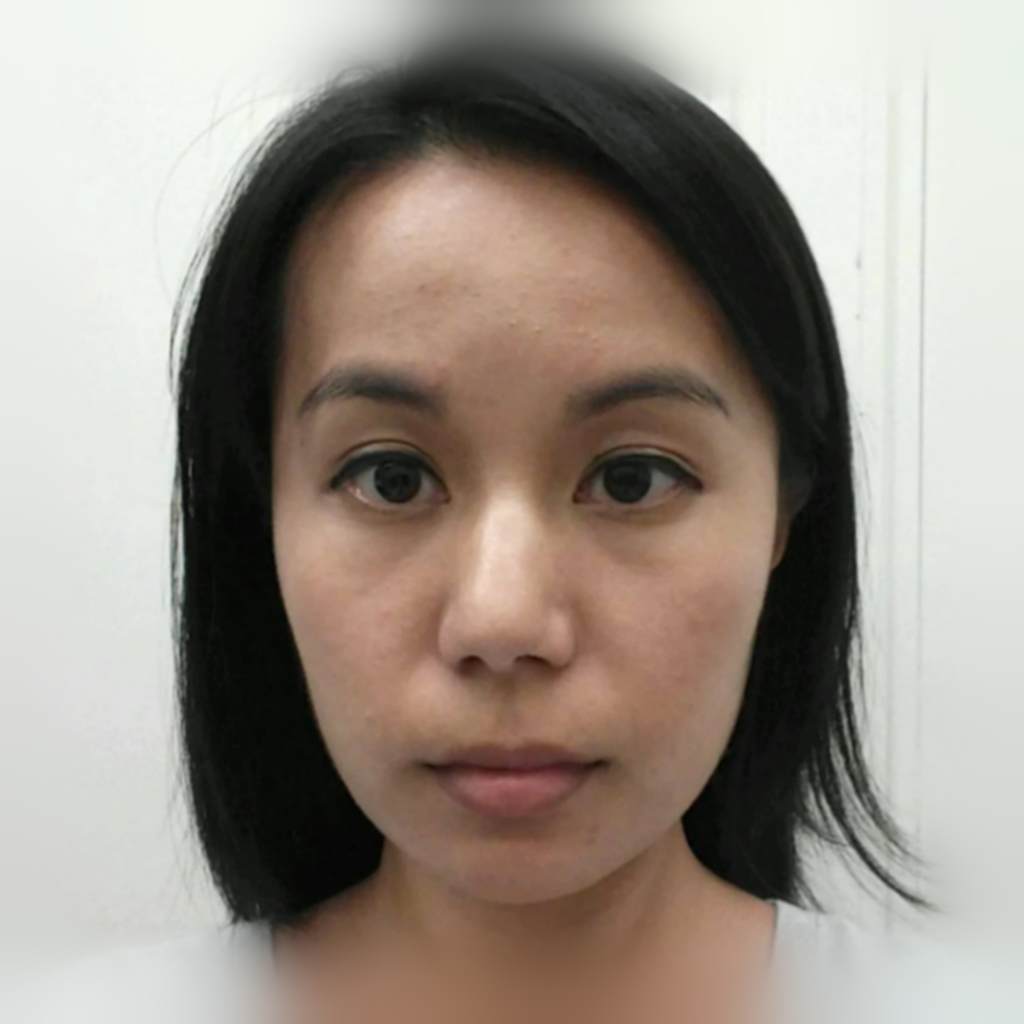}\includegraphics[width=0.2\linewidth]{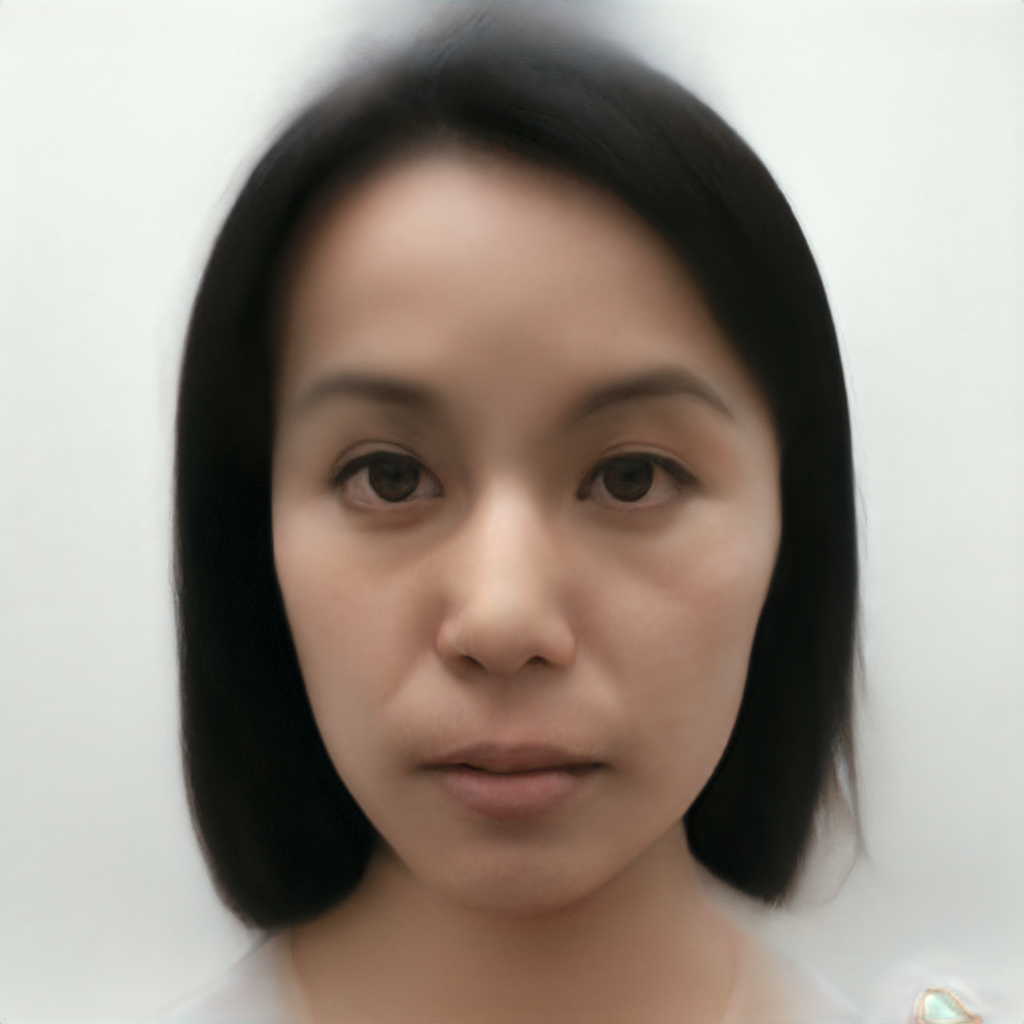}\includegraphics[width=0.2\linewidth]{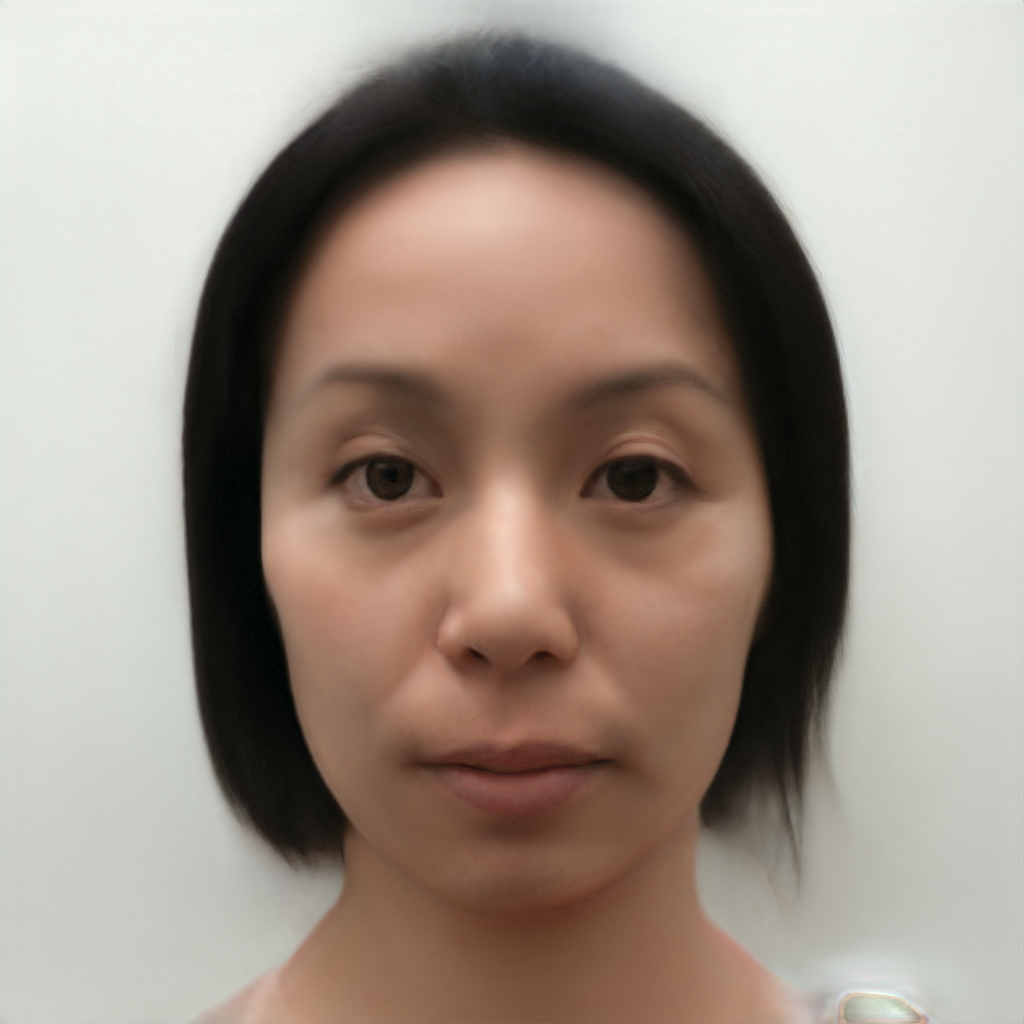}\includegraphics[width=0.2\linewidth]{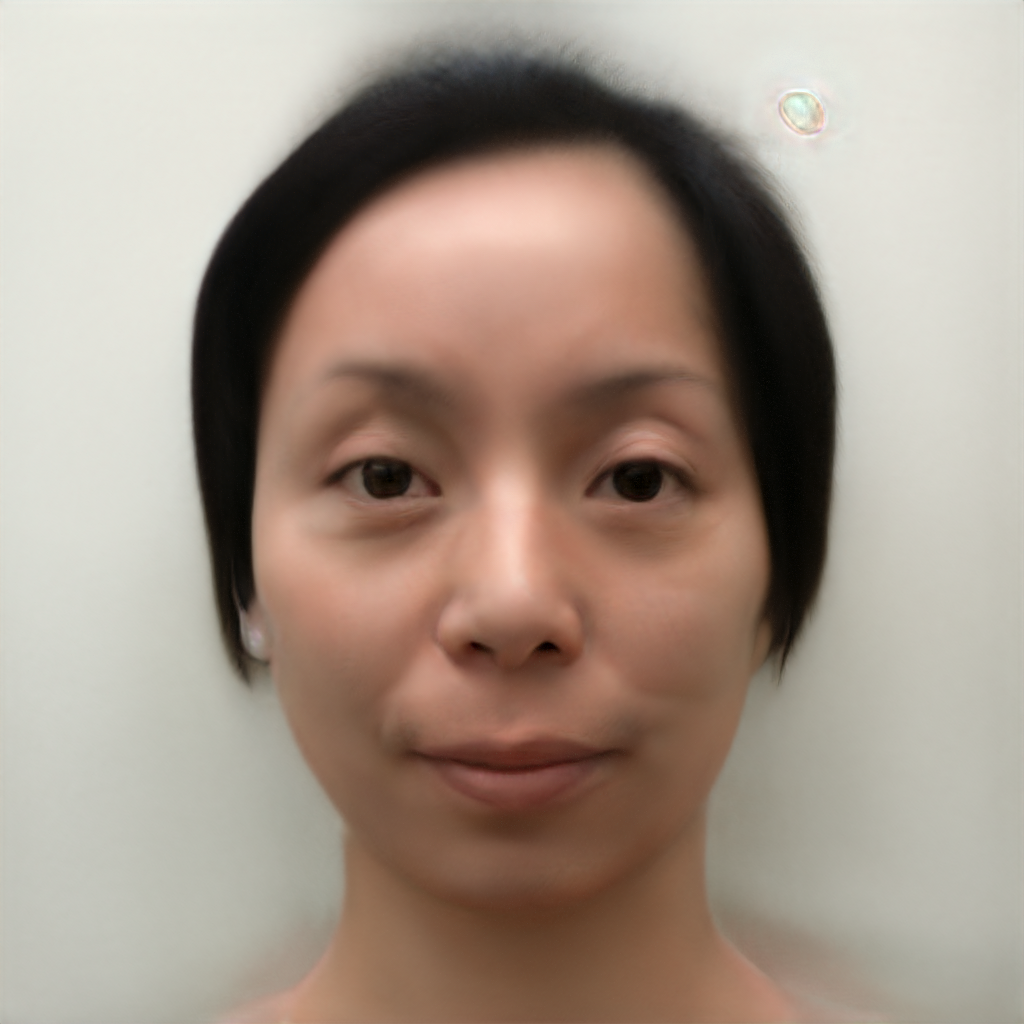}\includegraphics[width=0.2\linewidth]{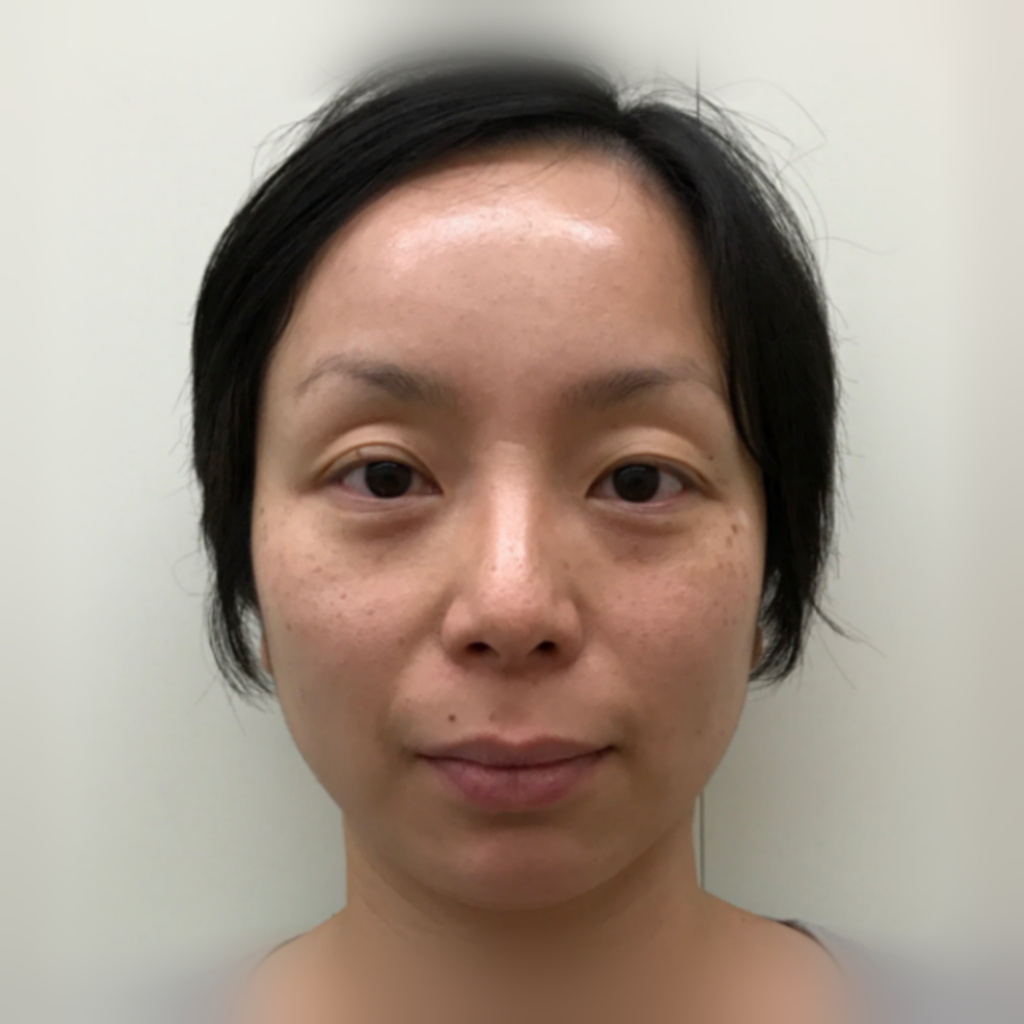}
   \includegraphics[width=0.2\linewidth]{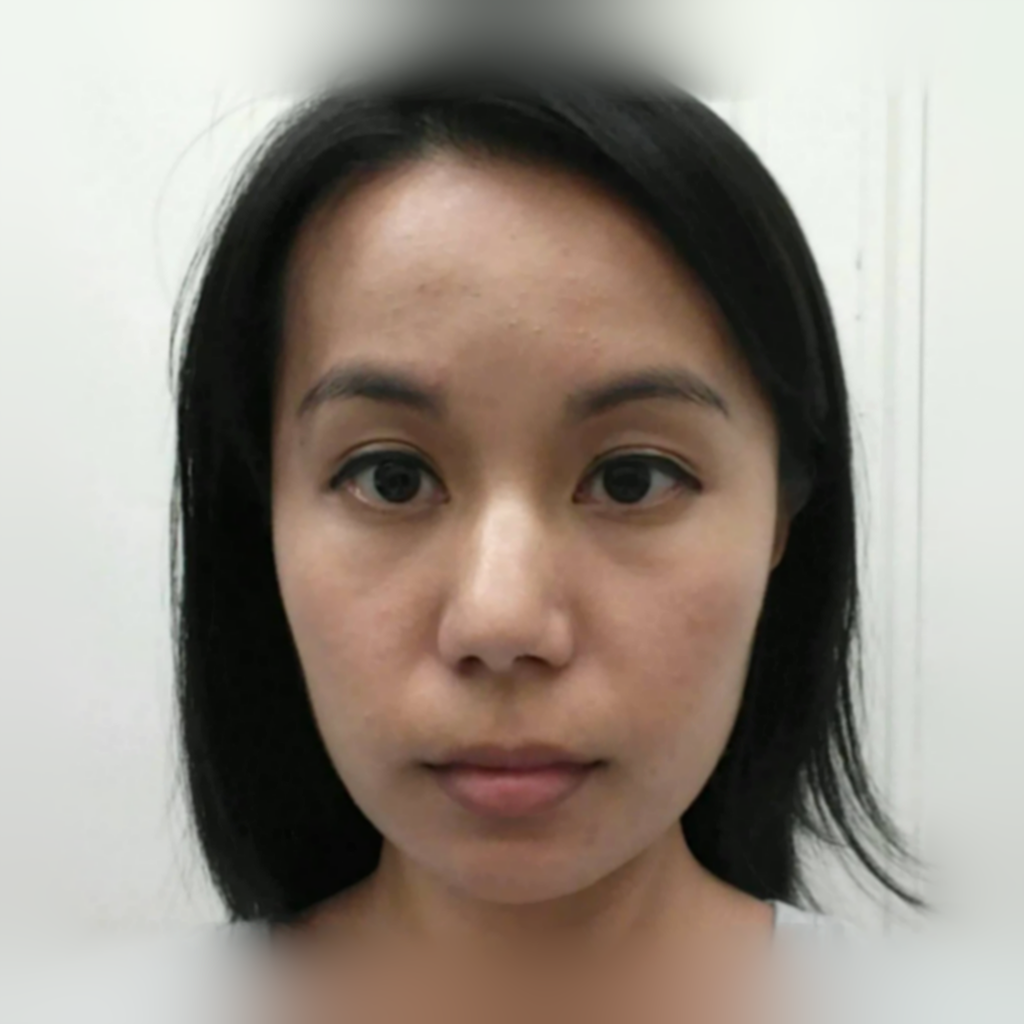}\includegraphics[width=0.2\linewidth]{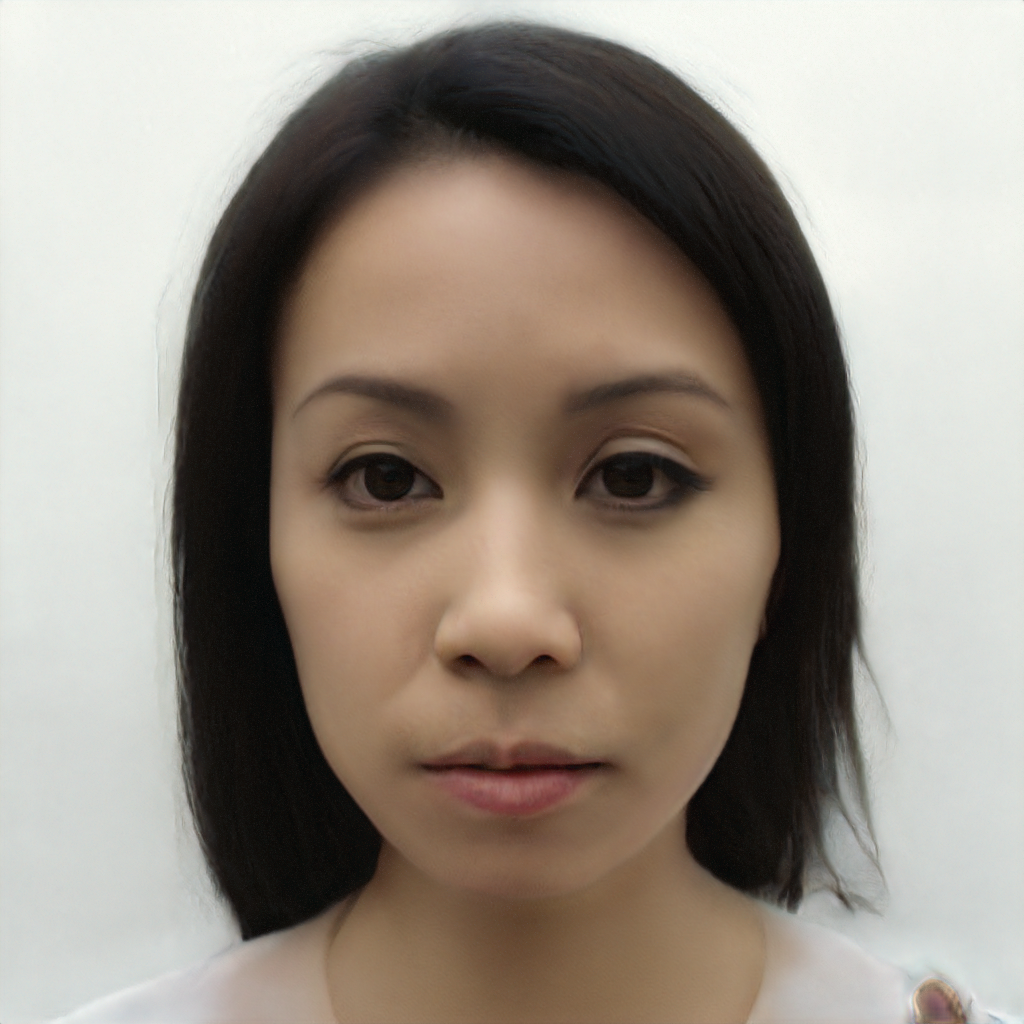}\includegraphics[width=0.2\linewidth]{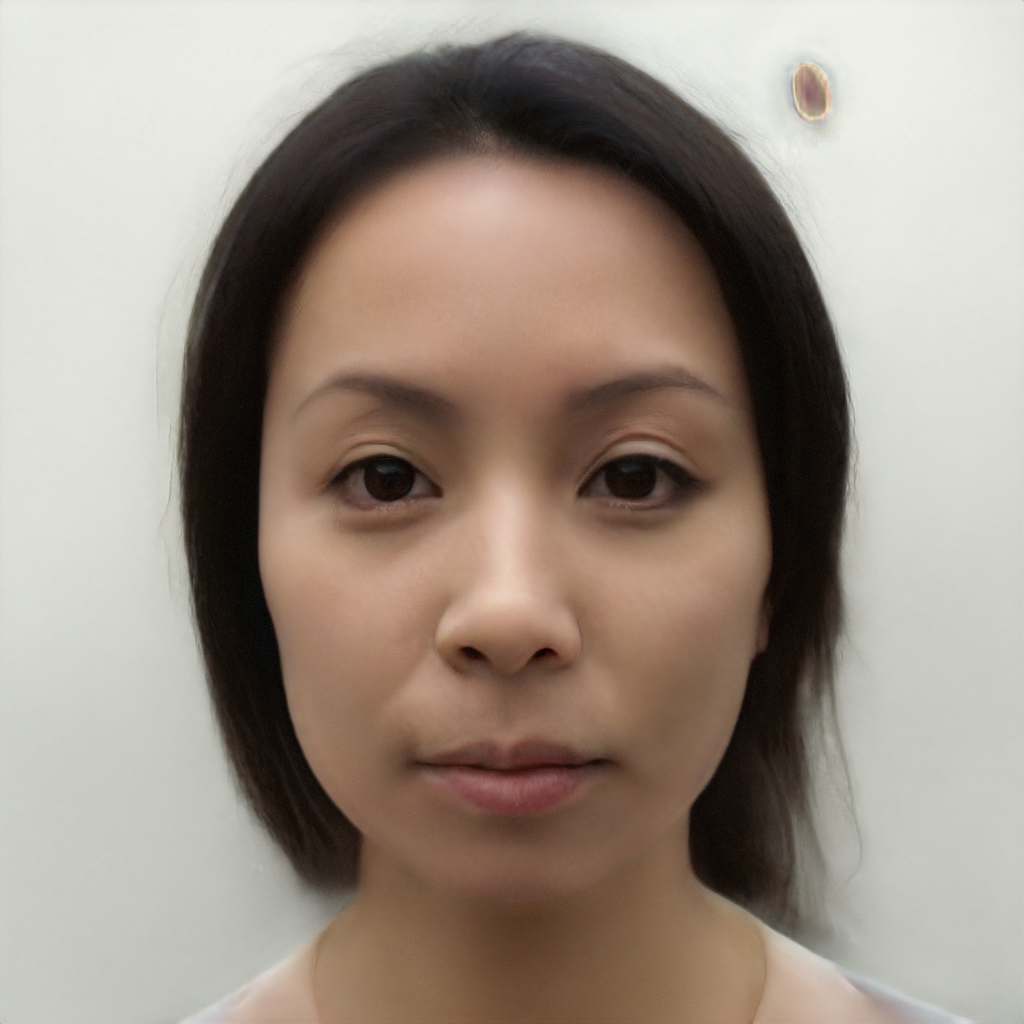}\includegraphics[width=0.2\linewidth]{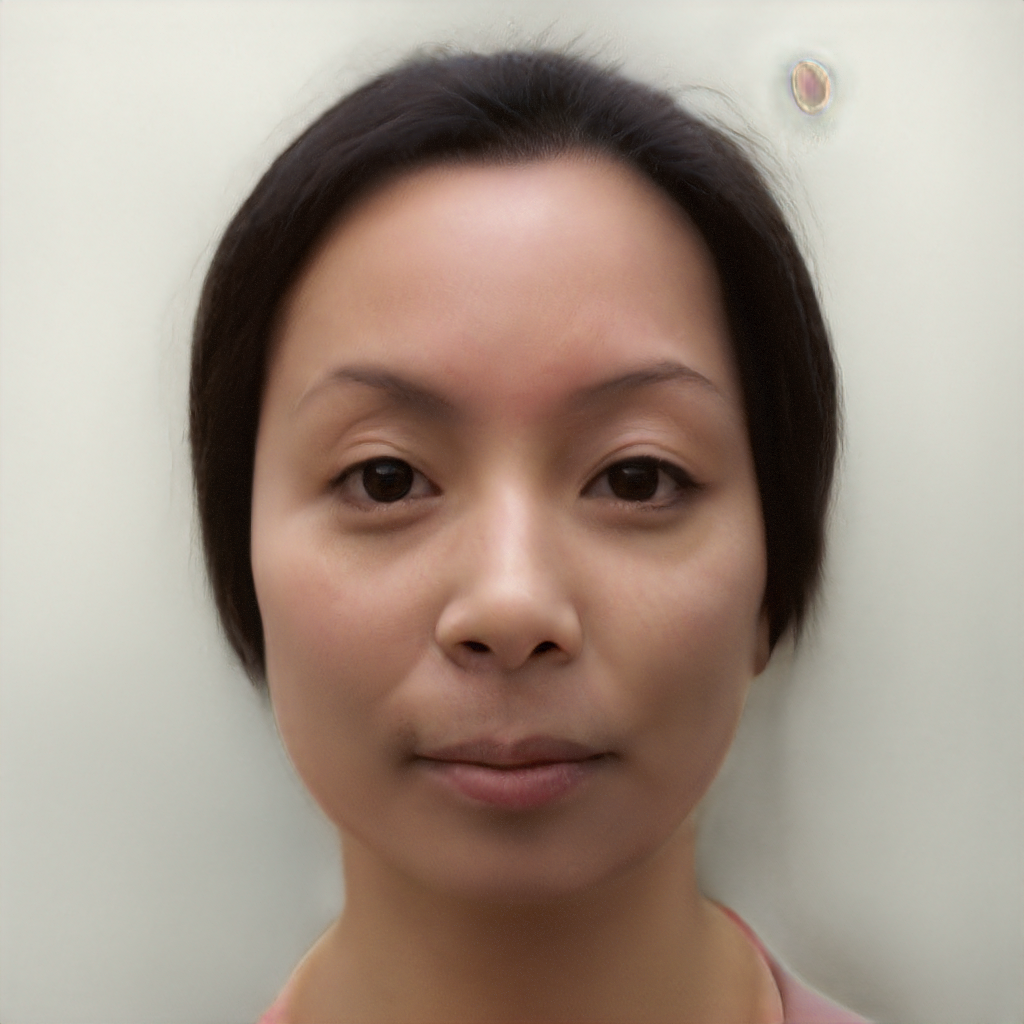}\includegraphics[width=0.2\linewidth]{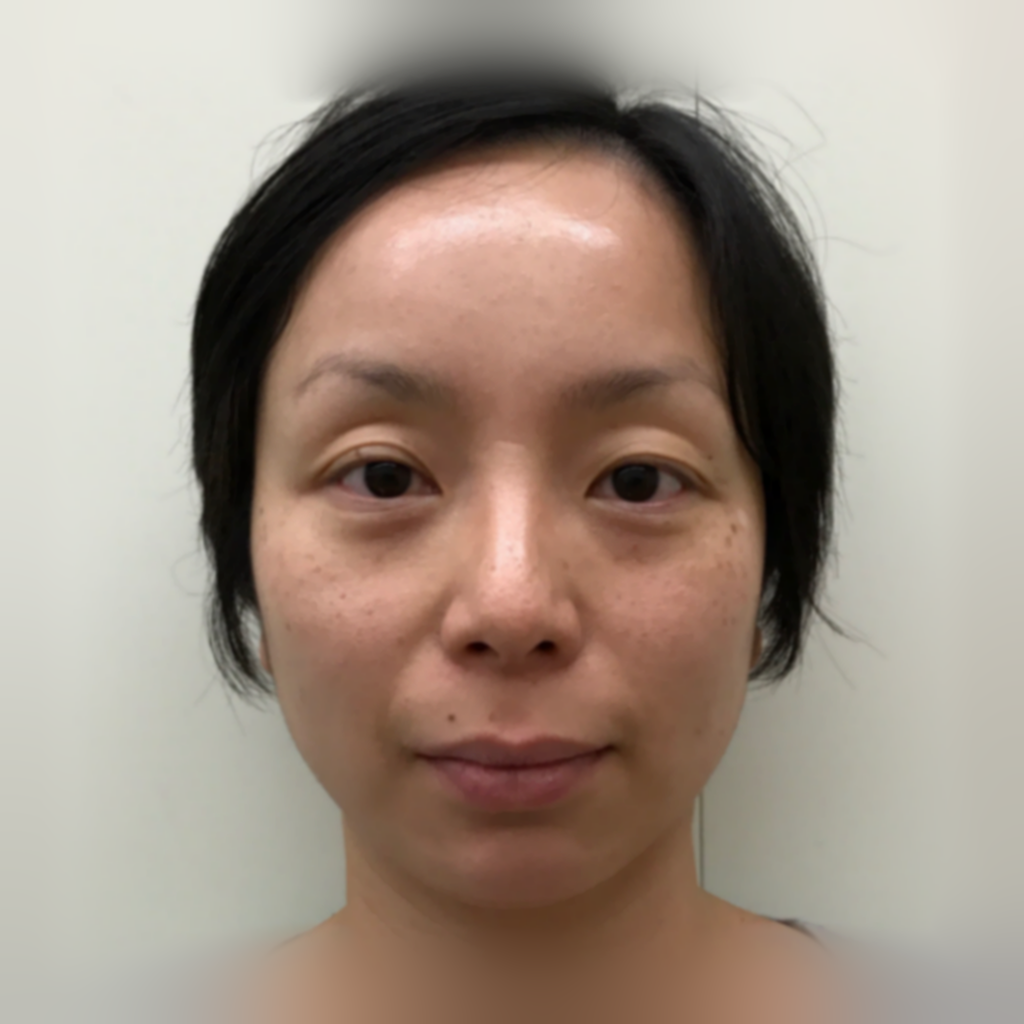}
\end{center}
   \caption{Ablation tests of the midpoint morphing method. Top - results of the full method as described in Section \ref{SubSec. METHOD Midpoint}; middle - perceptual loss and regularisation of the latent vector removed (i.e. reconstruction of pixel intensities only); bottom - use of non-independent latent vectors at each convolutional layer. (The full loss was used, as in the top row.)}
\label{fig:NIST Ablation}
\end{figure}

\section{Related Work}\label{Sec. Related Work}

\subsection{Securing systems against morphing attacks}

The largest part of the face-morphing attack literature consists of the development of methods for the detection of morphs, for example, by using deep learning techniques \cite{DBLP:conf/btas/DamerS0K18}, analysis of sensor noise in images \cite{DBLP:conf/iwbf/DebiasiSRUB18}, detection of landmark shifts \cite{DBLP:conf/icisp/ScherhagBGB18}, verification of the consistency of lighting conditions \cite{DBLP:conf/eusipco/SeiboldHE18}, or by de-morphing images to reveal the original subject \cite{DBLP:conf/eusipco/FerraraFM18}. Most current detection methods, however, are ineffective and suffer from high error rates that worsen when morphed images are printed and scanned \cite{DBLP:conf/eusipco/MakrushinW18, DBLP:conf/iwbf/ScherhagRRGRB17}. In \cite{DBLP:conf/eusipco/MakrushinW18} it is therefore recommended that identity document-issuing authorities enforce the submission of high-resolution digital images. However, they also point out that attackers could still manipulate digital noise signatures to obfuscate traces of image editing. In a recent survey of morphing attacks and detection methods \cite{DBLP:journals/access/ScherhagRMBB19} it was concluded that morphing attack detection methods do not generalise well to datasets incorporating real-world capture conditions. Indeed, in the most recent FRVT morph detection report \cite{ngan2020face}, the best value of APCER@BPCER=$0.01$ (Attack Presentation Classification Error Rate at a \textit{Bona fide} Presentation Classification Error Rate of $0.01$) for detection of morphs of the types shown in Figure \ref{fig:NIST Auto methods} (i.e. the ``Local Morph Colorized Match'', ``Splicing'', ``Combined'' and ``DST'' methods) was $88\%$ for the ``Splicing'' method.

An assessment of the vulnerability of FR to the \textit{average} of images of two identities \cite{DBLP:conf/icb/RaghavendraRVB17} showed it to be a more effective method than morphing. They also showed, however, that the averaged images were much easier to detect. It is unlikely, therefore, that an attacker would choose this type of method. In this work we propose two StyleGAN-based \textit{morphing} methods and, in light of the evident difficulty of detecting morphs, we instead focus on demonstrating the effect on morphing attacks of improvements to the robustness of FR algorithms.

\subsection{The development of style-based face-morphing}

\begin{figure}[t]
\begin{center}
   \includegraphics[width=1.0\linewidth]{plots/Header.png}
   \includegraphics[width=0.2\linewidth]{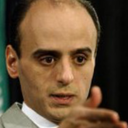}\includegraphics[width=0.2\linewidth]{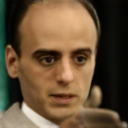}\includegraphics[width=0.2\linewidth]{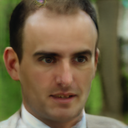}\includegraphics[width=0.2\linewidth]{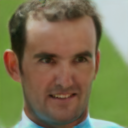}\includegraphics[width=0.2\linewidth]{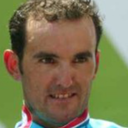}
   \includegraphics[width=0.2\linewidth]{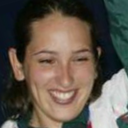}\includegraphics[width=0.2\linewidth]{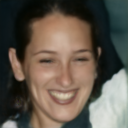}\includegraphics[width=0.2\linewidth]{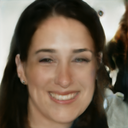}\includegraphics[width=0.2\linewidth]{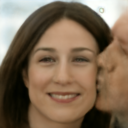}\includegraphics[width=0.2\linewidth]{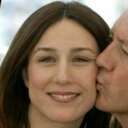}
   \includegraphics[width=0.2\linewidth]{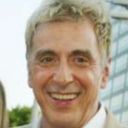}\includegraphics[width=0.2\linewidth]{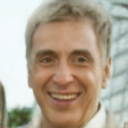}\includegraphics[width=0.2\linewidth]{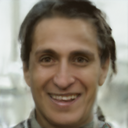}\includegraphics[width=0.2\linewidth]{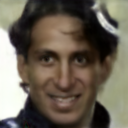}\includegraphics[width=0.2\linewidth]{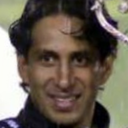}
   \includegraphics[width=0.2\linewidth]{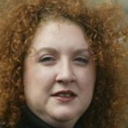}\includegraphics[width=0.2\linewidth]{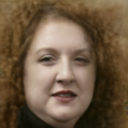}\includegraphics[width=0.2\linewidth]{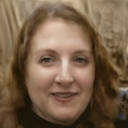}\includegraphics[width=0.2\linewidth]{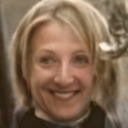}\includegraphics[width=0.2\linewidth]{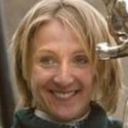}
   \includegraphics[width=0.2\linewidth]{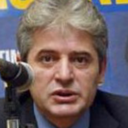}\includegraphics[width=0.2\linewidth]{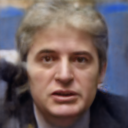}\includegraphics[width=0.2\linewidth]{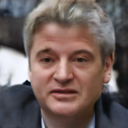}\includegraphics[width=0.2\linewidth]{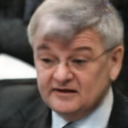}\includegraphics[width=0.2\linewidth]{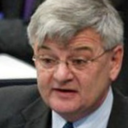}
\end{center}
   \caption{Example sets of image-reconstructions and morphs produced using the midpoint method. Each row represents a successful attack against Algo. 2017 but \textit{not} Algo. 2019 with acceptance thresholds set at FRR=0.25\%.}
\label{fig:Midpoint_examples_improved}
\end{figure}

Generative Adversarial Networks (GANs) \cite{DBLP:conf/nips/GoodfellowPMXWOCB14} learn to map latent vectors of random values to points on a manifold in data-space, usually image-space, representing realistic data-samples that can fool a concurrently trained discriminator into classifying them as real samples. Typically, generator architectures take a similar form to other deep neural networks, starting with the input - in this case a random vector - and applying a series of convolutions. In \cite{DBLP:conf/cvpr/KarrasLA19}, however, the vector of random values is projected to each convolutional layer of the network and used to directly influence the scale of variation in each feature map of each convolutional layer. This is achieved via conditional instance normalisation \cite{DBLP:conf/iclr/DumoulinSK17}, which was originally introduced as a method to manipulate the styles of images via the use of image-to-image translation networks \cite{DBLP:conf/cvpr/IsolaZZE17}. Since the generators of GANs do not translate images but \textit{grow} them, each convolutional layer naturally learns to control image features at different scales. For example, pose is controlled by early, large-scale features at low resolutions, whereas high-frequency, small-scale details such as wrinkles are controlled later on, at higher resolutions \cite{DBLP:conf/cvpr/KarrasLA19}. This natural, scale-wise disentanglement (in conjunction with the ``style mixing'' used in \cite{DBLP:conf/cvpr/KarrasLA19}) causes the projections of the latent vector - the so called ``$w^+$'' vectors - to be largely independent of one another. It is this independence, and therefore flexibility, that makes style-based GANs particularly suitable for image-reconstruction and then morphing.

\begin{figure}[t]
\begin{center}
   \includegraphics[width=1.0\linewidth]{plots/Header.png}
   \includegraphics[width=0.2\linewidth]{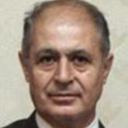}\includegraphics[width=0.2\linewidth]{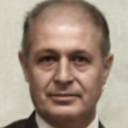}\includegraphics[width=0.2\linewidth]{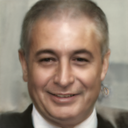}\includegraphics[width=0.2\linewidth]{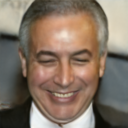}\includegraphics[width=0.2\linewidth]{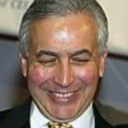}
   \includegraphics[width=0.2\linewidth]{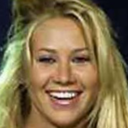}\includegraphics[width=0.2\linewidth]{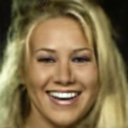}\includegraphics[width=0.2\linewidth]{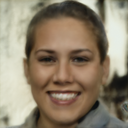}\includegraphics[width=0.2\linewidth]{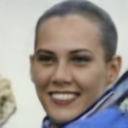}\includegraphics[width=0.2\linewidth]{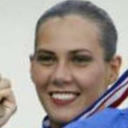}
   \includegraphics[width=0.2\linewidth]{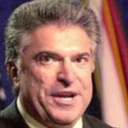}\includegraphics[width=0.2\linewidth]{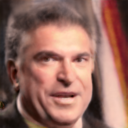}\includegraphics[width=0.2\linewidth]{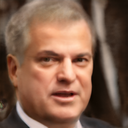}\includegraphics[width=0.2\linewidth]{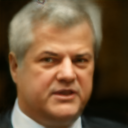}\includegraphics[width=0.2\linewidth]{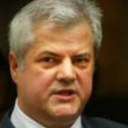}
   \includegraphics[width=0.2\linewidth]{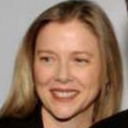}\includegraphics[width=0.2\linewidth]{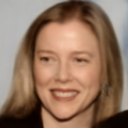}\includegraphics[width=0.2\linewidth]{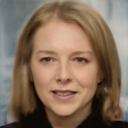}\includegraphics[width=0.2\linewidth]{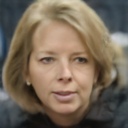}\includegraphics[width=0.2\linewidth]{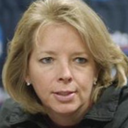}
   \includegraphics[width=0.2\linewidth]{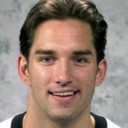}\includegraphics[width=0.2\linewidth]{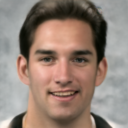}\includegraphics[width=0.2\linewidth]{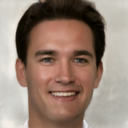}\includegraphics[width=0.2\linewidth]{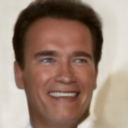}\includegraphics[width=0.2\linewidth]{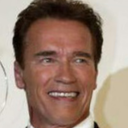}
\end{center}
   \caption{Example sets of image-reconstructions and morphs produced using the midpoint method. Each row represents a successful attack against both Algo. 2017 and Algo. 2019 with acceptance thresholds set at FRR=0.25\%.}
\label{fig:Midpoint_examples_problematic}
\end{figure}

In order to use GANs to perform morphing attacks, one needs to be able to \textit{invert} the generator, i.e. to find the latent vector that best describes some input image. There are various one-shot ways to do this, for example, one could train an encoder to regress the latents from synthetic images or, alternatively, train an encoder via Adversarially Learned Inference (ALI) \cite{DBLP:conf/iclr/DumoulinBPLAMC17, DBLP:conf/iclr/DonahueKD17} as was done in \cite{DBLP:conf/btas/DamerS0K18}. However, it is more effective, albeit slower, to find the latents using some iterative gradient descent method. Typically, it is difficult to fit GANs to non-synthetic images; so much so that the failure to reproduce images precisely has been used as a evidence that memorisation of images is not taking place in GANs \cite{DBLP:conf/cvpr/WebsterRSJ19}. However, in \cite{DBLP:conf/iccv/AbdalQW19} it was noticed that precise reconstructions could be achieved by treating the projected latents of StyleGAN independently during fitting, thereby taking advantage of the aforementioned scale-wise disentanglement. (This increase in precision can be seen by comparing the reconstructed images in the top and bottom rows of Figure \ref{fig:NIST Ablation}.) It is then straightforward to generate realistic face-morphs by linearly interpolating between two sets of recovered $w^+$ vectors. In this work, we also manipulate images in the $w^+$ latent space. Further details of our methods are given in the following section.

\section{Face-morphing with StyleGAN}\label{Sec. Face-morphing with StyleGAN}

We will evaluate robustness of FR algorithms to two different methods of face-morphing based on StyleGAN: the ``midpoint method'', which is similar to that demonstrated in \cite{DBLP:conf/iccv/AbdalQW19} and \cite{DBLP:conf/iwbf/VenkateshZRRDB20}, and the ``dual biometric method'', which was developed for this study. In both methods we optimise the loss functions using Adam \cite{DBLP:journals/corr/KingmaB14}. To speed up convergence and improve reconstruction quality, initialisations of the latent vectors are provided by a one-shot encoder trained on pairs of random vectors and associated synthetic images.

\subsection{The midpoint method}\label{SubSec. METHOD Midpoint}

\begin{figure*}[t]
    \centering
    \includegraphics[width=0.49\linewidth]{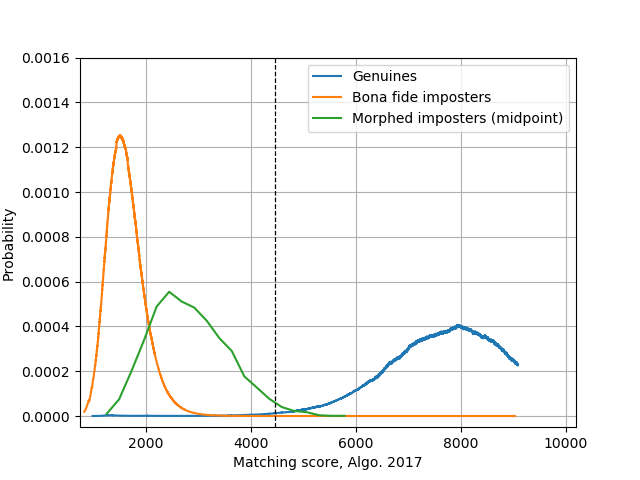}\includegraphics[width=0.49\linewidth]{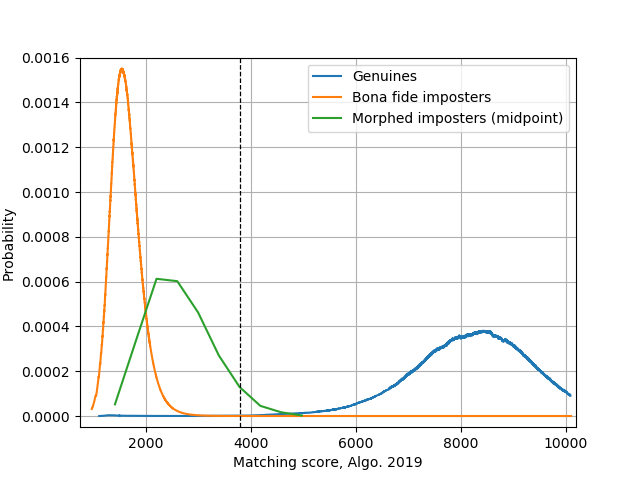} \\
    \includegraphics[width=0.49\linewidth]{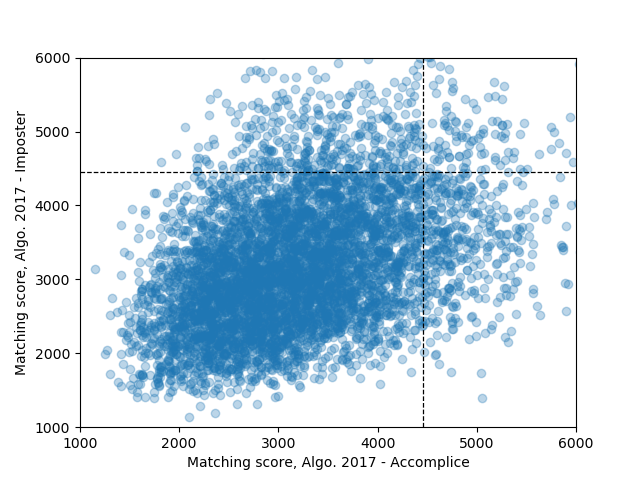}\includegraphics[width=0.49\linewidth]{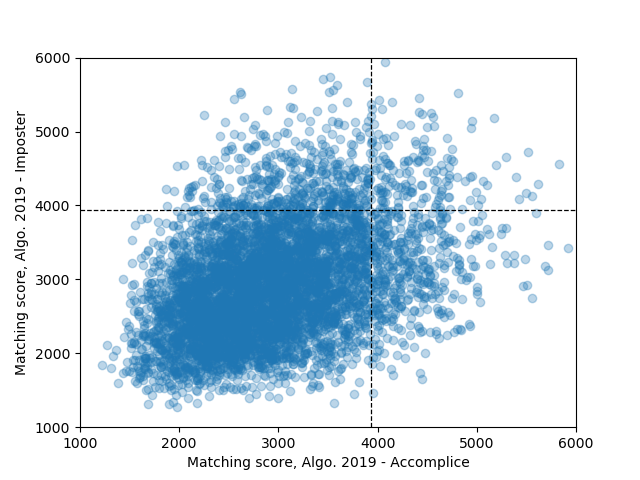}
    \caption{Distributions of matching scores for Algo. 2017 (left) and Algo. 2019 (right). Morphed imposters were produced using the \textbf{midpoint method}. Dashed lines represent thresholds of FAR=$1\times10^{-5}$ for \textit{bona fide} imposters.}
    \label{fig:Midpoint_OLD_NEW}
\end{figure*}

Face-morphing using the midpoint method consists of two steps: recovering the two latent vectors that best describe two input images, and generating a synthetic image from the midpoint interpolation of those two vectors. To recover $\mathbf{w^+}$ for an input image $\mathbf{x}$, the following loss function is minimised:
\begin{align}
  \mathcal{L}_{w^+} \label{eq. midpoint} = &\mathcal{P}(G(\mathbf{w^+}),\mathbf{x}) + \frac{\lambda_r}{N_x} ||G(\mathbf{w^+}) - \mathbf{x}||^2_2 \nonumber \\
  &+ \lambda_w ||\mathbf{w^+} - \mathbf{\overline{w}}||_1
\end{align}
where $G$ is the generator (with StyleGAN's mapping network removed), $N_x$ is the number of image pixels, $\mathbf{\overline{w}}$ is the average of the $\mathbf{w^+}$ seen during training of $G$, and $\mathcal{P}$ is a perceptual loss given by
\begin{align}
  \mathcal{P}(G(\mathbf{w^+})),\mathbf{x}) \label{eq. perceptual} = &\frac{\lambda_v}{N_v} ||VGG_9(G(\mathbf{w^+})-VGG_9(\mathbf{x})||^2_2 \nonumber \\
  &+ \lambda_m (1 - MSSSIM(G(\mathbf{w^+}), \mathbf{x}))
\end{align}
where $VGG_9$ is the output of the ninth layer of the VGG classification network \cite{DBLP:journals/corr/SimonyanZ14a} used to extract discriminative features, $N_v$ is the number of VGG features, and $MSSSIM$ is the Tensorflow implementation of the MS-SSIM metric described in \cite{1292216}. The generator, $G$, is the official version of the StyleGAN generator trained on the Flickr-Faces-HQ dataset. The code implementing the inversion of StyleGAN's generator was taken from \cite{styleganencoder} and the coefficients weighting each term of equations (\ref{eq. midpoint}) and (\ref{eq. perceptual}) are left at their default values of $\lambda_r = 1.5$, $\lambda_w = 0.5$, $\lambda_v = 0.4$ and $\lambda_m = 200$. Once two vectors, $\mathbf{w}^+_1$ and $\mathbf{w}^+_2$ have been recovered, the final morphed image is given by
\begin{align}
  \mathbf{x}_{morph} = G(\frac{\mathbf{w}^+_1 + \mathbf{w}^+_2}{2})
\end{align}
In the middle row of Figure \ref{fig:NIST Ablation} we demonstrate the effect of reverting the method to that used in \cite{DBLP:conf/iccv/AbdalQW19} by removing the perceptual loss term and the regularisation of $\mathbf{w^+}$ from equation (\ref{eq. midpoint}) during latent recovery. The reconstructed images as well the midpoint morph become more blurred, lacking in high-frequency detail. This result motivates our use of the full, perceptual loss function in our experiments. Results showing the level of robustness of FR to morphing attacks using the full midpoint method are given in Section \ref{SubSec. RESULTS Midpoint}.

\subsection{The dual biometric method}\label{SubSec. METHOD Biometric}

Although the latent space of StyleGAN is disentangled with respect to some scale-dependent features, identity features are not necessarily disentangled. This means that the equality
\begin{align}
  B(\mathbf{x}_{morph}) = \frac{1}{2}B(G(\mathbf{w}^+_1)) + \frac{1}{2}B(G(\mathbf{w}^+_2))
\end{align}
where $B$ is a biometric network producing an identity feature vector, does not necessarily hold; i.e. the identity of the midpoint morph does not necessarily lie between the identities of the two reconstructed images in a biometric feature space. A more reliable method of ensuring that the morphed identity remains close to each of the original identities could be to explicitly minimise those distances in the feature space. This motivates our dual biometric method in which the following cost function is minimised:
\begin{align}
  \mathcal{L}_{w^+} \label{eq. bio} = &||B(G(\mathbf{w^+}))-B(\mathbf{x_1})||^2_2 \nonumber \\
  &+ ||B(G(\mathbf{w^+}))-B(\mathbf{x_2})||^2_2 \nonumber \\
  &+ \lambda_w ||\mathbf{w^+} - \mathbf{\overline{w}}||_1
\end{align}
For $B$ we used a Keras implementation of the VGGFace2 ``SENet'' network \cite{DBLP:conf/fgr/CaoSXPZ18} taken from \cite{kerasvggface}. We have also included the same L1 regularisation of the latent vector as was used in the midpoint method. Since the biometric loss terms are robust to (i.e. ignore) all image features except for the identity, the L1 regularisation is important for maintaining a realistic looking image. $\lambda_w$ was tuned by hand based on the appearance of a handful of morphed images and set to a value of $3$. Results showing the level of robustness of FR to morphing attacks using this second method are given in Section \ref{SubSec. RESULTS Biometric}.

\section{Results}\label{Sec. Results}

We evaluate robustness of FR to morphing attacks using the Labeled Faces in the Wild (LFW) dataset \cite{LFWTech}. To simulate realistic morphing attacks we first select the highest quality image for each of the $5478$ identities. We then assign fifty random ``friends'' to each identity and select the strongest identity match to be the accomplice, as judged by a biometric matching algorithm. Morphed images were produced for each of these image pairs using both the midpoint method and our dual biometric method. The original, \textit{bona fide} images were then matched against their mated morphs. Note that we do not compare morphs with independent images of the mated subjects. Comparisons are made with the \textit{bona fide} images used to create the morphs meaning that matching scores are likely to be at a maximum, thus giving conservative estimates of FR system vulnerability. We present results for two matching algorithms, the first based on DeepVisage \cite{DBLP:conf/iccvw/HasnatBMG017} that we will refer to as ``Algo. 2017'' and the second based on ArcFace \cite{DBLP:conf/cvpr/DengGXZ19} that we will refer to as ``Algo. 2019''.

\subsection{Results - The midpoint method}\label{SubSec. RESULTS Midpoint}

Figures \ref{fig:Midpoint_examples_improved} and \ref{fig:Midpoint_examples_problematic} give examples of face-morphs generated from pairs from LWF using the midpoint method. In Figure \ref{fig:Midpoint_OLD_NEW} we plot distributions of matching scores produced for this type of morph by the 2017 and 2019 algorithms. The green, ``Morphed imposters'' curves show the distributions of the minimum mated morph similarity scores (MMMSS), i.e. the minimum of either the accomplice-morph or morph-imposter matching score. The minimum score is interesting since is the strength of the weakest similarity that determines whether the attack as a whole succeeds. The blue, ``Genuines'' curves show the distribution of mated matching scores for sets of \textit{bona fide} images from LFW sharing the same identity, and the orange, ``Bona fide imposters'' curves show non-mated matching scores. In each figure we have drawn a threshold at the score corresponding to a False Acceptance Rate (FAR) of $1\times 10^{-5}$ based on the distribution of \textit{bona fide} imposters. Values of Mated Morph Presentation Match Rate (MMPMR) \cite{DBLP:conf/biosig/ScherhagNRGVSSM17} at this threshold are presented in Table \ref{tab:MMPMR} and correspond to the proportions of points lying in the top-right quadrant of the scatter plots.

In typical circumstances, both algorithms are able to well separate the distributions of mated and non-mated matching scores. However, inclusion of minimum mated morph similarity scores (MMMSSs) blurs this separation and at FAR=$1\times 10^{-5}$ the success rate of simulated morphing attacks, the MMPMR, is $1.99\%$ for Algo. 2017 and $2.96\%$ for Algo. 2019, i.e. the MMPMR is three orders of magnitude larger than the FAR. What is more, despite the Genuine and Imposter curves clearly being better separated by Algo. 2019, the value of MMPMR@FAR=$1\times10^{-5}$ increases. This is not because MMMSSs become less distinguishable from \textit{bona fide} mated matching scores; in fact, from Figure \ref{fig:Midpoint_OLD_NEW} we also see an improvement in separation of the Genuine and Morphed imposter curves. The issue arises from the fact that the improvement in separation of the \textit{bona fide} mated and non-mated matching scores is larger than for the MMMSSs. This acts to shift the threshold of FAR=$10^{-5}$ to a lower score that ``overtakes'' the improvements in MMMSS. This means that we cannot assume that improvements to FR systems will lead directly to increased robustness to morphing attacks. Instead, the response of the FR system to datasets of morphed images should be considered when setting operational thresholds.

\begin{table}[t]
\begin{center}
\begin{tabular}{|l|c|c|}
\hline
Morphing method & Algo. 2017 & Algo. 2019 \\
\hline\hline
Genuines (FRR) & 0.73\% & 0.25\% \\
Midpoint Morphs (MMPMR) & 1.99\% & 2.96\% \\
Biometric Morphs (MMPMR) & 3.88\% & 2.34\% \\
\hline
\end{tabular}
\end{center}
\caption{MMPMRs and FRR at a False Acceptance Rate of $1\times 10^{-5}$ for two different face-recognition algorithms.}
\label{tab:MMPMR}
\end{table}

\begin{figure}[t]
    \centering
    \includegraphics[width=1.0\linewidth]{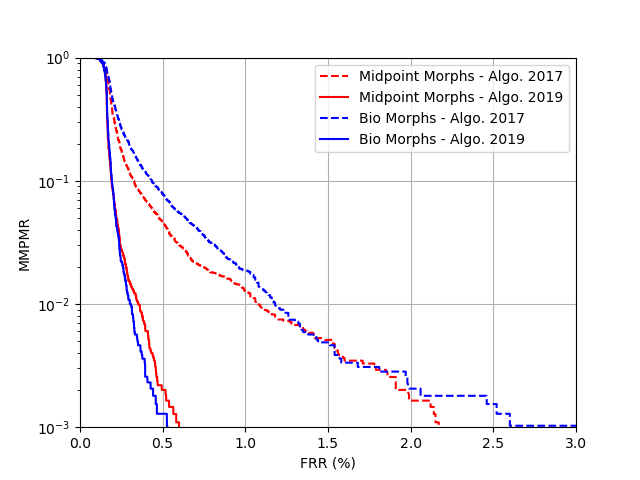}
    \caption{ROC curves showing the trade-off between MMPMR and FRR.}
    \label{fig:MMPMR_FRR_ROC}
\end{figure}

Table \ref{tab:Fixed_FRR} shows values of MMPMR, and also FAR, at a threshold corresponding to FRR=$0.73\%$ for the \textit{bona fide} mated pairs of LFW. We see that this corresponds to the original threshold of FAR=$1\times10^{-5}$ for Algo. 2017 but that for Algo. 2019 it corresponds to FAR=$1.81\times10^{-6}$. At this much more stringent threshold, MMPMR drops to $0.07\%$ for Algo. 2019 (and drops to $0\%$ for the morphs produced by the dual biometric method). This means that, by compromising on improvements to FRR, essentially \textit{all} face-morphing attacks of the type presented here can be prevented.  \cite{DBLP:conf/biosig/ScherhagNRGVSSM17} suggests reporting Relative Morph Match Rate (RMMR) as a measure of FR system vulnerability where RMMR = MMPMR + FRR. This measure varies with threshold, however, and implicitly weights robustness to morphs and low FRR as being equal in priority, which is not necessarily the case. We find it preferable to observe the compromise between MMPMR and FRR by plotting the relevant ROC curve, as has been done in Figure \ref{fig:MMPMR_FRR_ROC}. Here we see that the ROC curves for Algo. 2019 are significantly steeper than for Algo. 2017 indicating that accepting only a small increase in FRR can cause the success rates of morphing attacks to plummet relative to those measured against Algo. 2017.

\begin{table}[t]
\begin{center}
\begin{tabular}{|l|c|c|}
\hline
Morphing method & Algo. 2017 & Algo. 2019 \\
\hline\hline
\textit{Bone fide} imposters (FAR) & $1\times10^{-5}$ & $1.81\times10^{-6}$ \\
Midpoint Morphs (MMPMR) & 1.99\% & 0.07\% \\
Biometric Morphs (MMPMR) & 3.88\% & 0.00\% \\
\hline
\end{tabular}
\end{center}
\caption{MMPMRs and FAR at a False Rejection Rate of $0.73\%$ for two different face-recognition algorithms.}
\label{tab:Fixed_FRR}
\end{table}

\begin{figure}[t]
\begin{center}
   \includegraphics[width=0.8\linewidth]{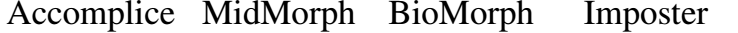}
   \includegraphics[width=0.2\linewidth]{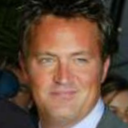}\includegraphics[width=0.2\linewidth]{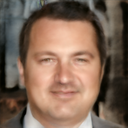}\includegraphics[width=0.2\linewidth]{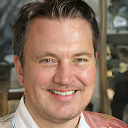}\includegraphics[width=0.2\linewidth]{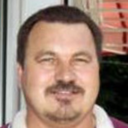}
   \includegraphics[width=0.2\linewidth]{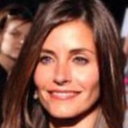}\includegraphics[width=0.2\linewidth]{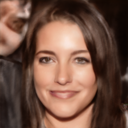}\includegraphics[width=0.2\linewidth]{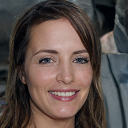}\includegraphics[width=0.2\linewidth]{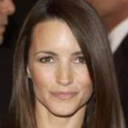}
   \includegraphics[width=0.2\linewidth]{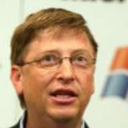}\includegraphics[width=0.2\linewidth]{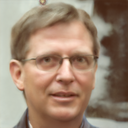}\includegraphics[width=0.2\linewidth]{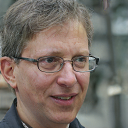}\includegraphics[width=0.2\linewidth]{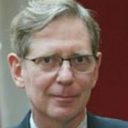}
   \includegraphics[width=0.2\linewidth]{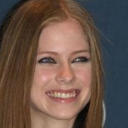}\includegraphics[width=0.2\linewidth]{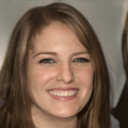}\includegraphics[width=0.2\linewidth]{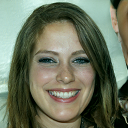}\includegraphics[width=0.2\linewidth]{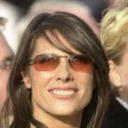}
\end{center}
   \caption{Comparison of morphs generated using the midpoint and dual biometric methods.}
\label{fig:Mid_Bio_comp}
\end{figure}

\begin{figure*}[t]
    \centering
    \includegraphics[width=0.49\linewidth]{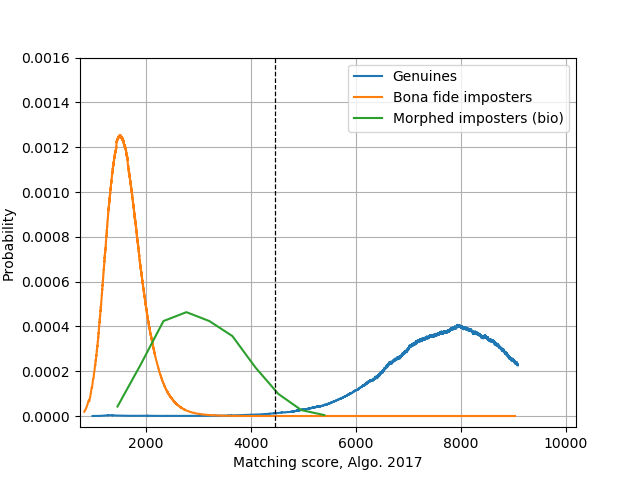}\includegraphics[width=0.49\linewidth]{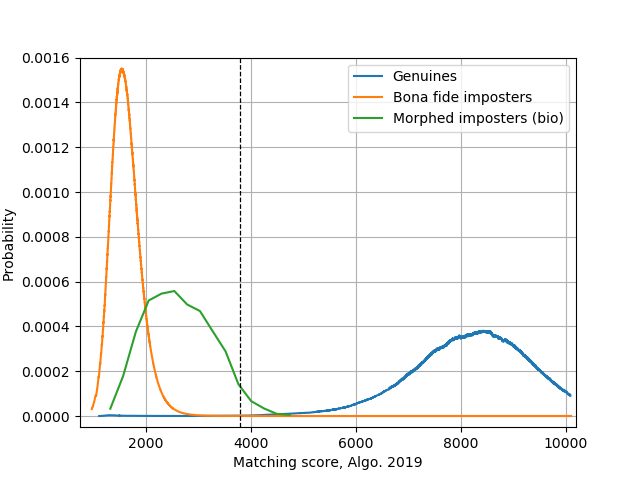} \\
    \includegraphics[width=0.49\linewidth]{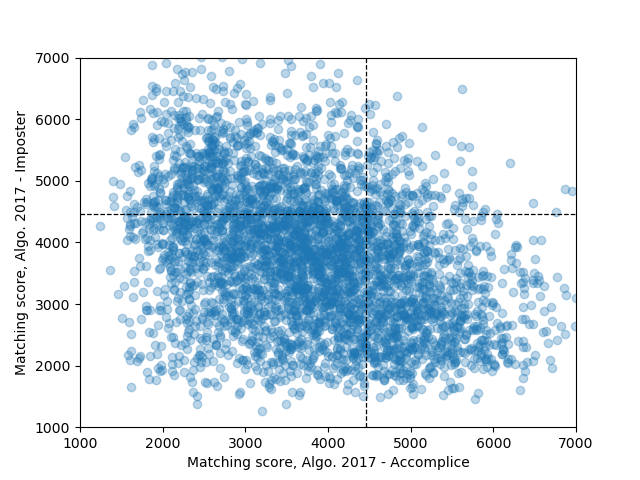}\includegraphics[width=0.49\linewidth]{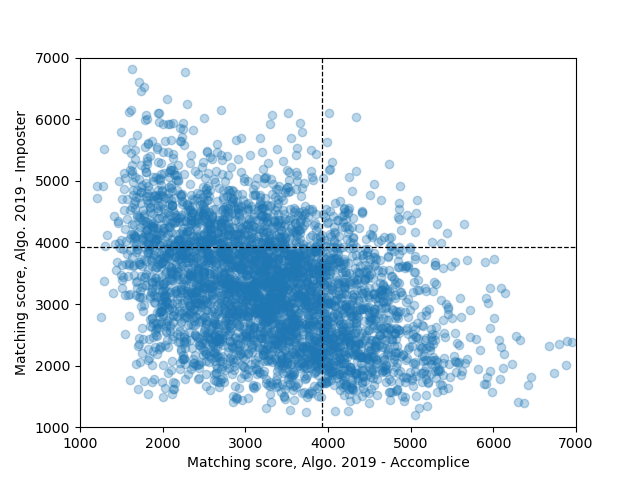}
    \caption{Distributions of matching scores for Algo. 2017 (left) and Algo. 2019 (right). Morphed imposters were produced using the \textbf{dual biometric method}. Dashed lines represent thresholds of FAR=$1\times10^{-5}$ for \textit{bona fide} imposters.}
    \label{fig:Bio_OLD_NEW}
\end{figure*}

\subsection{Results - The dual biometric method}\label{SubSec. RESULTS Biometric}

Figure \ref{fig:Mid_Bio_comp} gives a direct comparison of face-morphs generated using the dual biometric method (``BioMorph'') with those generated by the midpoint method (``MidMorph''). We see that the biometrically morphed identities are plausible and that they are distinct from the midpoint morphs. We also notice that the image-quality of the bio-morphs is higher than that of the midpoint morphs. This is because the images of LFW are of a lower quality than the images of FFHQ used to train StyleGAN. Image artefacts from LFW therefore seep into the midpoint morphs via the image-reconstructions. During generation of the bio-morphs, the original images are never reconstructed. The only constraint is that similar identity-related features be generated.

\begin{figure}[t]
\begin{center}
   \includegraphics[width=0.6\linewidth]{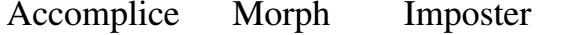}
   \includegraphics[width=0.2\linewidth]{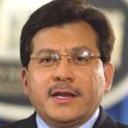}\includegraphics[width=0.2\linewidth]{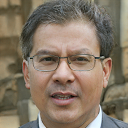}\includegraphics[width=0.2\linewidth]{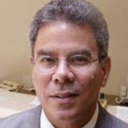}
   \includegraphics[width=0.2\linewidth]{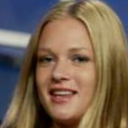}\includegraphics[width=0.2\linewidth]{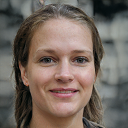}\includegraphics[width=0.2\linewidth]{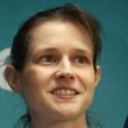}
   \includegraphics[width=0.2\linewidth]{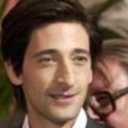}\includegraphics[width=0.2\linewidth]{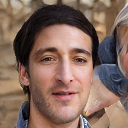}\includegraphics[width=0.2\linewidth]{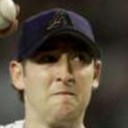}
   \includegraphics[width=0.2\linewidth]{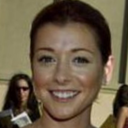}\includegraphics[width=0.2\linewidth]{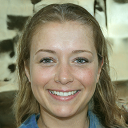}\includegraphics[width=0.2\linewidth]{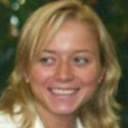}
   \includegraphics[width=0.2\linewidth]{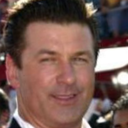}\includegraphics[width=0.2\linewidth]{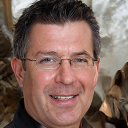}\includegraphics[width=0.2\linewidth]{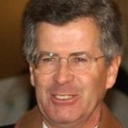}
\end{center}
   \caption{Examples of morphs produced using the dual biometric method. Each row represents a successful attack against Algo. 2017 but \textit{not} Algo. 2019 with acceptance thresholds set at FRR=0.25\%.}
\label{fig:Bio_examples_improved}
\end{figure}

\begin{figure}[t]
\begin{center}
   \includegraphics[width=0.6\linewidth]{plots/Header_bio.png}
   \includegraphics[width=0.2\linewidth]{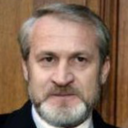}\includegraphics[width=0.2\linewidth]{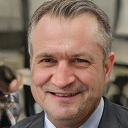}\includegraphics[width=0.2\linewidth]{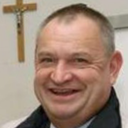}
   \includegraphics[width=0.2\linewidth]{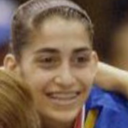}\includegraphics[width=0.2\linewidth]{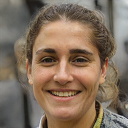}\includegraphics[width=0.2\linewidth]{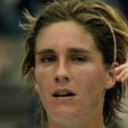}
   \includegraphics[width=0.2\linewidth]{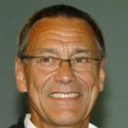}\includegraphics[width=0.2\linewidth]{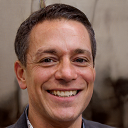}\includegraphics[width=0.2\linewidth]{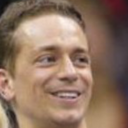}
   \includegraphics[width=0.2\linewidth]{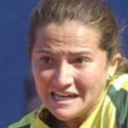}\includegraphics[width=0.2\linewidth]{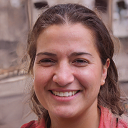}\includegraphics[width=0.2\linewidth]{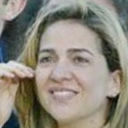}
   \includegraphics[width=0.2\linewidth]{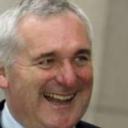}\includegraphics[width=0.2\linewidth]{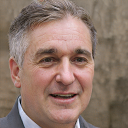}\includegraphics[width=0.2\linewidth]{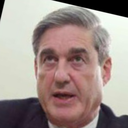}
\end{center}
   \caption{Examples of morphs produced using the dual biometric method. Each row represents a successful attack against both Algo. 2017 and Algo. 2019 with acceptance thresholds set at FRR=0.25\%.}
\label{fig:Bio_examples_problematic}
\end{figure}

Figure \ref{fig:Bio_OLD_NEW} shows the distributions of matching scores for the biometrically morphed images. From the scatter plots, we see that the matching scores are less balanced between accomplices and imposters, i.e. a large proportion of bio morphs were found to match one identity much more strongly than the other. This contrasts with the midpoint method where a larger proportion of morphs were found to give weak matching scores for both original images. From Table \ref{tab:MMPMR} we see that despite the imbalanced matching scores, MMPMR@FAR=$1\times 10^{-5}$ is larger for biometric morphs than for midpoint morphs as measured by Algo. 2017. This situation reverses, however, for Algo. 2019 which succeeds in reducing the number of successful simulated attacks without modification to the threshold of FAR=$1\times 10^{-5}$. Table \ref{tab:Fixed_FRR} shows that by sacrificing improvements to FRR and modifying the acceptance threshold of Algo. 2019 to FAR=$1.81\times 10^{-6}$, all simulated biometric morphing attacks can be prevented. Figure \ref{fig:Bio_examples_problematic} shows examples of bio-morphs that remain to be problematic for the 2019 algorithm with a threshold at FAR=$1\times 10^{-5}$. Those shown in Figure \ref{fig:Bio_examples_improved} were previously problematic for Algo. 2017 but are correctly rejected by Algo. 2019.

\begin{figure}[t]
    \centering
    \includegraphics[width=0.25\linewidth]{plots/NIST_subjectA.png}\includegraphics[width=0.25\linewidth]{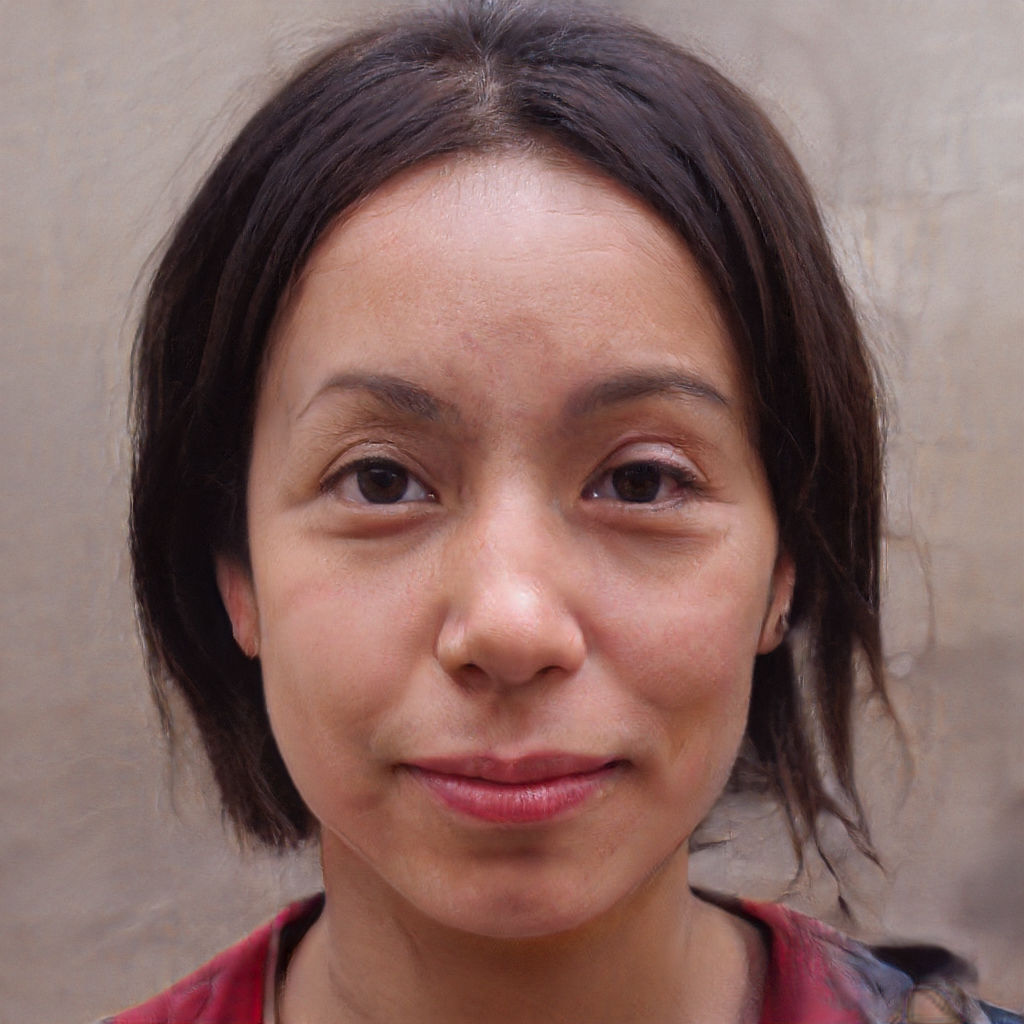}\includegraphics[width=0.25\linewidth]{plots/NIST_subjectB.png}
    \includegraphics[width=0.25\linewidth]{plots/NIST_subjectA.png}\includegraphics[width=0.25\linewidth]{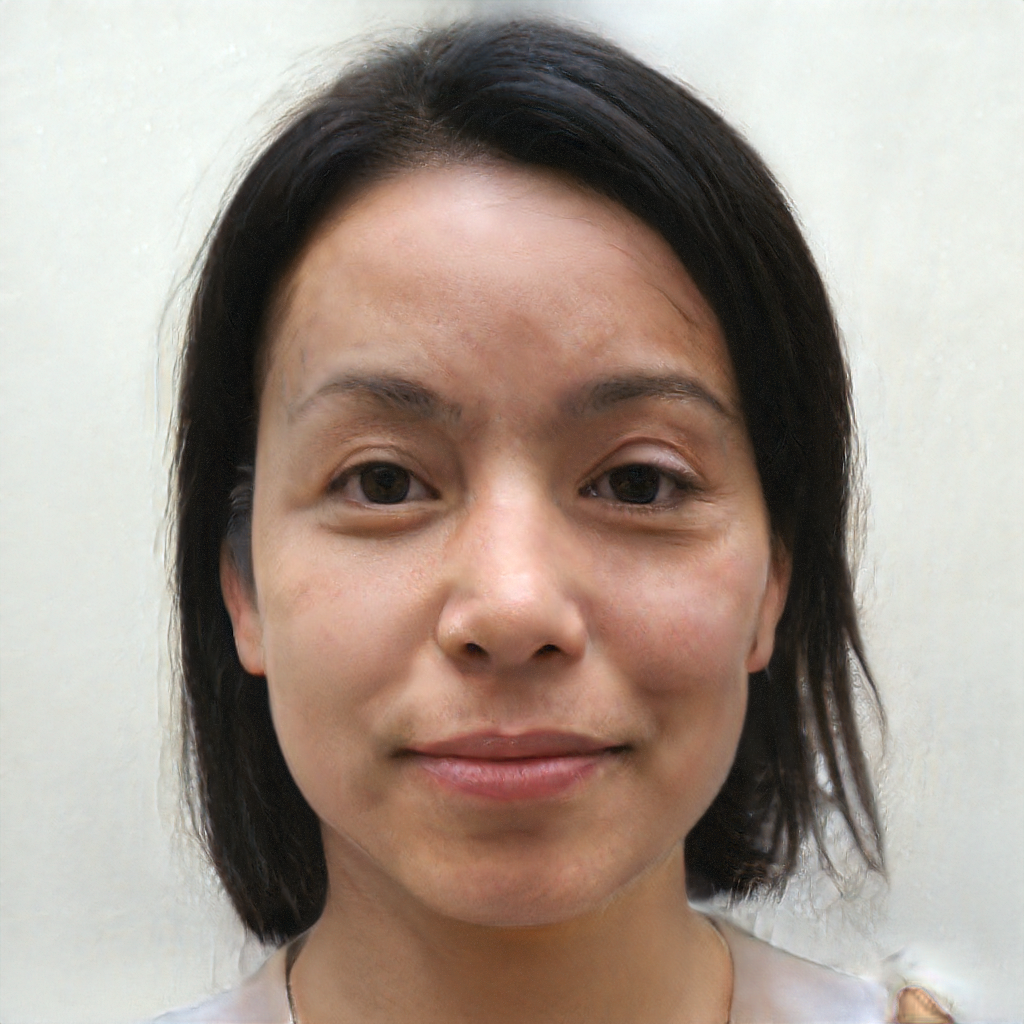}\includegraphics[width=0.25\linewidth]{plots/NIST_subjectB.png}
    \caption{Demonstration of the dual biometric method applied to the passport-style images of Figure \ref{fig:NIST Morph}. Top: as described in equation (\ref{eq. bio}); Bottom: with an added reconstruction loss on the background regions.}
    \label{fig:NIST Bio Morph}
\end{figure}

By evaluating the dual biometric method on an in-the-wild dataset, we have inadvertently disguised the fact that desirable (for the attacker), non-identity image characteristics are lost. For example, Figure \ref{fig:NIST Bio Morph} (top) demonstrates the result of applying the biometric morphing method to the passport-style photographs of Figure \ref{fig:NIST Morph}. The generated morph resembles an in-the-wild image from FFHQ and would likely not be accepted for use on an identity document. To avoid this problem, a pixel-wise reconstruction loss can be applied to background regions. Figure \ref{fig:NIST Bio Morph} (bottom) shows an example of a biometric morph produced in this way. An alternative approach for an attacker to overcome this issue could be to train a GAN using solely passport-style images.

\section{Conclusions}\label{Sec. Conclusions}

In this work we evaluated the robustness of two facial recognition algorithms to face-morphing attacks using a recent, StyleGAN-based morphing method. We also proposed and evaluated a second, related morphing method in which distances of morphed faces from the contributing identities in a biometric feature space are minimised explicitly. Both morphing methods were found to be of potential threat with their relative success rates depending on the particular FR algorithm under attack. Assuming that we have been able to simulate realistic attack scenarios, it is likely that fewer than $3\%$ of StyleGAN-based morphing attacks would succeed against a state-of-the-art facial recognition algorithm with a matching threshold set at FAR=$1\times 10^{-5}$. We also observed that improvements to FR algorithms do not necessarily translate directly to increased robustness to face-morphing attacks and recommend that matching scores for datasets of morphed images be considered when setting operational acceptance thresholds.

{\small
\bibliographystyle{ieee_fullname}
\bibliography{egbib}

\begin{thebibliography}{1}\itemsep=-1pt

\bibitem{Alpher02}
FirstName Alpher.
\newblock Frobnication.
\newblock {\em IEEE Trans. Pattern Anal. Mach. Intell.}, 12(1):234--778, 2002.

\bibitem{Alpher03}
FirstName Alpher and FirstName Fotheringham-Smythe.
\newblock Frobnication revisited.
\newblock {\em Journal of Foo}, 13(1):234--778, 2003.

\bibitem{Alpher04}
FirstName Alpher, FirstName Fotheringham-Smythe, and FirstName Gamow.
\newblock Can a machine frobnicate?
\newblock {\em Journal of Foo}, 14(1):234--778, 2004.

\bibitem{Authors14}
FirstName LastName.
\newblock The frobnicatable foo filter, 2014.
\newblock Face and Gesture submission ID 324. Supplied as additional material
  {\tt fg324.pdf}.

\bibitem{Authors14b}
FirstName LastName.
\newblock Frobnication tutorial, 2014.
\newblock Supplied as additional material {\tt tr.pdf}.

\end{thebibliography}


\begin{thebibliography}{10}\itemsep=-1pt

\bibitem{DBLP:conf/iccv/AbdalQW19}
Rameen Abdal, Yipeng Qin, and Peter Wonka.
\newblock Image2stylegan: How to embed images into the stylegan latent space?
\newblock In {\em 2019 {IEEE/CVF} International Conference on Computer Vision,
  {ICCV} 2019, Seoul, Korea (South), October 27 - November 2, 2019}, pages
  4431--4440. {IEEE}, 2019.

\bibitem{styleganencoder}
Peter Baylies.
\newblock stylegan-encoder.
\newblock
  \url{https://github.com/pbaylies/stylegan-encoder/tree/266d1eb2da09894adcb49879fbd0674e12cab739},
  2019.

\bibitem{DBLP:conf/fgr/CaoSXPZ18}
Qiong Cao, Li Shen, Weidi Xie, Omkar~M. Parkhi, and Andrew Zisserman.
\newblock Vggface2: {A} dataset for recognising faces across pose and age.
\newblock In {\em 13th {IEEE} International Conference on Automatic Face {\&}
  Gesture Recognition, {FG} 2018, Xi'an, China, May 15-19, 2018}, pages 67--74.
  {IEEE} Computer Society, 2018.

\bibitem{DBLP:conf/btas/DamerS0K18}
Naser Damer, Alexandra~Mosegui Saladie, Andreas Braun, and Arjan Kuijper.
\newblock Morgan: Recognition vulnerability and attack detectability of face
  morphing attacks created by generative adversarial network.
\newblock In {\em 9th {IEEE} International Conference on Biometrics Theory,
  Applications and Systems, {BTAS} 2018, Redondo Beach, CA, USA, October 22-25,
  2018}, pages 1--10. {IEEE}, 2018.

\bibitem{DBLP:conf/iwbf/DebiasiSRUB18}
Luca Debiasi, Ulrich Scherhag, Christian Rathgeb, Andreas Uhl, and Christoph
  Busch.
\newblock Prnu-based detection of morphed face images.
\newblock In {\em 2018 International Workshop on Biometrics and Forensics,
  {IWBF} 2018, Sassari, Italy, June 7-8, 2018}, pages 1--7. {IEEE}, 2018.

\bibitem{DBLP:conf/cvpr/DengGXZ19}
Jiankang Deng, Jia Guo, Niannan Xue, and Stefanos Zafeiriou.
\newblock Arcface: Additive angular margin loss for deep face recognition.
\newblock In {\em {IEEE} Conference on Computer Vision and Pattern Recognition,
  {CVPR} 2019, Long Beach, CA, USA, June 16-20, 2019}, pages 4690--4699.
  Computer Vision Foundation / {IEEE}, 2019.

\bibitem{DBLP:conf/iclr/DonahueKD17}
Jeff Donahue, Philipp Kr{\"{a}}henb{\"{u}}hl, and Trevor Darrell.
\newblock Adversarial feature learning.
\newblock In {\em 5th International Conference on Learning Representations,
  {ICLR} 2017, Toulon, France, April 24-26, 2017, Conference Track
  Proceedings}. OpenReview.net, 2017.

\bibitem{DBLP:conf/iclr/DumoulinBPLAMC17}
Vincent Dumoulin, Ishmael Belghazi, Ben Poole, Alex Lamb, Mart{\'{\i}}n
  Arjovsky, Olivier Mastropietro, and Aaron~C. Courville.
\newblock Adversarially learned inference.
\newblock In {\em 5th International Conference on Learning Representations,
  {ICLR} 2017, Toulon, France, April 24-26, 2017, Conference Track
  Proceedings}. OpenReview.net, 2017.

\bibitem{DBLP:conf/iclr/DumoulinSK17}
Vincent Dumoulin, Jonathon Shlens, and Manjunath Kudlur.
\newblock A learned representation for artistic style.
\newblock In {\em 5th International Conference on Learning Representations,
  {ICLR} 2017, Toulon, France, April 24-26, 2017, Conference Track
  Proceedings}. OpenReview.net, 2017.

\bibitem{DBLP:conf/icb/FerraraFM14}
Matteo Ferrara, Annalisa Franco, and Davide Maltoni.
\newblock The magic passport.
\newblock In {\em {IEEE} International Joint Conference on Biometrics,
  Clearwater, {IJCB} 2014, FL, USA, September 29 - October 2, 2014}, pages
  1--7. {IEEE}, 2014.

\bibitem{DBLP:conf/eusipco/FerraraFM18}
Matteo Ferrara, Annalisa Franco, and Davide Maltoni.
\newblock Face demorphing in the presence of facial appearance variations.
\newblock In {\em 26th European Signal Processing Conference, {EUSIPCO} 2018,
  Roma, Italy, September 3-7, 2018}, pages 2365--2369. {IEEE}, 2018.

\bibitem{DBLP:conf/nips/GoodfellowPMXWOCB14}
Ian~J. Goodfellow, Jean Pouget{-}Abadie, Mehdi Mirza, Bing Xu, David
  Warde{-}Farley, Sherjil Ozair, Aaron~C. Courville, and Yoshua Bengio.
\newblock Generative adversarial nets.
\newblock In Zoubin Ghahramani, Max Welling, Corinna Cortes, Neil~D. Lawrence,
  and Kilian~Q. Weinberger, editors, {\em Advances in Neural Information
  Processing Systems 27: Annual Conference on Neural Information Processing
  Systems 2014, December 8-13 2014, Montreal, Quebec, Canada}, pages
  2672--2680, 2014.

\bibitem{DBLP:conf/iccvw/HasnatBMG017}
Md.~Abul Hasnat, Julien Bohn{\'{e}}, Jonathan Milgram, St{\'{e}}phane Gentric,
  and Liming Chen.
\newblock Deepvisage: Making face recognition simple yet with powerful
  generalization skills.
\newblock In {\em 2017 {IEEE} International Conference on Computer Vision
  Workshops, {ICCV} Workshops 2017, Venice, Italy, October 22-29, 2017}, pages
  1682--1691. {IEEE} Computer Society, 2017.

\bibitem{LFWTech}
Gary~B. Huang, Manu Ramesh, Tamara Berg, and Erik Learned-Miller.
\newblock Labeled faces in the wild: A database for studying face recognition
  in unconstrained environments.
\newblock Technical Report 07-49, University of Massachusetts, Amherst, October
  2007.

\bibitem{DBLP:conf/cvpr/IsolaZZE17}
Phillip Isola, Jun{-}Yan Zhu, Tinghui Zhou, and Alexei~A. Efros.
\newblock Image-to-image translation with conditional adversarial networks.
\newblock In {\em 2017 {IEEE} Conference on Computer Vision and Pattern
  Recognition, {CVPR} 2017, Honolulu, HI, USA, July 21-26, 2017}, pages
  5967--5976. {IEEE} Computer Society, 2017.

\bibitem{DBLP:conf/cvpr/KarrasLA19}
Tero Karras, Samuli Laine, and Timo Aila.
\newblock A style-based generator architecture for generative adversarial
  networks.
\newblock In {\em {IEEE} Conference on Computer Vision and Pattern Recognition,
  {CVPR} 2019, Long Beach, CA, USA, June 16-20, 2019}, pages 4401--4410.
  Computer Vision Foundation / {IEEE}, 2019.

\bibitem{DBLP:journals/corr/KingmaB14}
Diederik~P. Kingma and Jimmy Ba.
\newblock Adam: {A} method for stochastic optimization.
\newblock In Yoshua Bengio and Yann LeCun, editors, {\em 3rd International
  Conference on Learning Representations, {ICLR} 2015, San Diego, CA, USA, May
  7-9, 2015, Conference Track Proceedings}, 2015.

\bibitem{DBLP:conf/eusipco/MahfoudiTRMDP19}
Ga{\"{e}}l Mahfoudi, Badr Tajini, Florent Retraint, Fr{\'{e}}d{\'{e}}ric
  Morain{-}Nicolier, Jean{-}Luc Dugelay, and Marc Pic.
\newblock {DEFACTO:} image and face manipulation dataset.
\newblock In {\em 27th European Signal Processing Conference, {EUSIPCO} 2019,
  {A} Coru{\~{n}}a, Spain, September 2-6, 2019}, pages 1--5. {IEEE}, 2019.

\bibitem{DBLP:conf/eusipco/MakrushinW18}
Andrey Makrushin and Andreas Wolf.
\newblock An overview of recent advances in assessing and mitigating the face
  morphing attack.
\newblock In {\em 26th European Signal Processing Conference, {EUSIPCO} 2018,
  Roma, Italy, September 3-7, 2018}, pages 1017--1021. {IEEE}, 2018.

\bibitem{kerasvggface}
Refik~Can Malli.
\newblock keras-vggface.
\newblock
  \url{https://github.com/rcmalli/keras-vggface/tree/9ac97da84f18e392f5009d83cafcc5359204a408},
  2020.

\bibitem{ngan2020face}
Mei Ngan, Patrick Grother, Kayee Hanaoka, and Jason Kuo.
\newblock Face recognition vendor test (frvt) part 4: Morph performance of
  automated face morph detection.
\newblock {\em National Institute of Technology (NIST), Tech. Rep. NISTIR},
  8292, 2020.

\bibitem{DBLP:conf/icb/RaghavendraRVB17}
Ramachandra Raghavendra, Kiran~B. Raja, Sushma Venkatesh, and Christoph Busch.
\newblock Face morphing versus face averaging: Vulnerability and detection.
\newblock In {\em 2017 {IEEE} International Joint Conference on Biometrics,
  {IJCB} 2017, Denver, CO, USA, October 1-4, 2017}, pages 555--563. {IEEE},
  2017.

\bibitem{DBLP:conf/icisp/ScherhagBGB18}
Ulrich Scherhag, Dhanesh Budhrani, Marta Gomez{-}Barrero, and Christoph Busch.
\newblock Detecting morphed face images using facial landmarks.
\newblock In Alamin Mansouri, Abderrahim Elmoataz, Fathallah Nouboud, and Driss
  Mammass, editors, {\em Image and Signal Processing - 8th International
  Conference, {ICISP} 2018, Cherbourg, France, July 2-4, 2018, Proceedings},
  volume 10884 of {\em Lecture Notes in Computer Science}, pages 444--452.
  Springer, 2018.

\bibitem{DBLP:conf/biosig/ScherhagNRGVSSM17}
Ulrich Scherhag, Andreas Nautsch, Christian Rathgeb, Marta Gomez{-}Barrero,
  Raymond N.~J. Veldhuis, Luuk~J. Spreeuwers, Maikel Schils, Davide Maltoni,
  Patrick Grother, S{\'{e}}bastien Marcel, Ralph Breithaupt, Ramachandra
  Raghavendra, and Christoph Busch.
\newblock Biometric systems under morphing attacks: Assessment of morphing
  techniques and vulnerability reporting.
\newblock In Arslan Br{\"{o}}mme, Christoph Busch, Antitza Dantcheva, Christian
  Rathgeb, and Andreas Uhl, editors, {\em International Conference of the
  Biometrics Special Interest Group, {BIOSIG} 2017, Darmstadt, Germany,
  September 20-22, 2017}, volume {P-270} of {\em {LNI}}, pages 149--159. {GI} /
  {IEEE}, 2017.

\bibitem{DBLP:conf/iwbf/ScherhagRRGRB17}
Ulrich Scherhag, Ramachandra Raghavendra, Kiran~B. Raja, Marta Gomez{-}Barrero,
  Christian Rathgeb, and Christoph Busch.
\newblock On the vulnerability of face recognition systems towards morphed face
  attacks.
\newblock In {\em 5th International Workshop on Biometrics and Forensics,
  {IWBF} 2017, Coventry, United Kingdom, April 4-5, 2017}, pages 1--6. {IEEE},
  2017.

\bibitem{DBLP:journals/access/ScherhagRMBB19}
Ulrich Scherhag, Christian Rathgeb, Johannes Merkle, Ralph Breithaupt, and
  Christoph Busch.
\newblock Face recognition systems under morphing attacks: {A} survey.
\newblock {\em {IEEE} Access}, 7:23012--23026, 2019.

\bibitem{DBLP:conf/eusipco/SeiboldHE18}
Clemens Seibold, Anna Hilsmann, and Peter Eisert.
\newblock Reflection analysis for face morphing attack detection.
\newblock In {\em 26th European Signal Processing Conference, {EUSIPCO} 2018,
  Roma, Italy, September 3-7, 2018}, pages 1022--1026. {IEEE}, 2018.

\bibitem{DBLP:journals/corr/SimonyanZ14a}
Karen Simonyan and Andrew Zisserman.
\newblock Very deep convolutional networks for large-scale image recognition.
\newblock In Yoshua Bengio and Yann LeCun, editors, {\em 3rd International
  Conference on Learning Representations, {ICLR} 2015, San Diego, CA, USA, May
  7-9, 2015, Conference Track Proceedings}, 2015.

\bibitem{DBLP:conf/iwbf/VenkateshZRRDB20}
Sushma Venkatesh, Haoyu Zhang, Raghavendra Ramachandra, Kiran~B. Raja, Naser
  Damer, and Christoph Busch.
\newblock Can {GAN} generated morphs threaten face recognition systems equally
  as landmark based morphs? - vulnerability and detection.
\newblock In {\em 8th International Workshop on Biometrics and Forensics,
  {IWBF} 2020, Porto, Portugal, April 29-30, 2020}, pages 1--6. {IEEE}, 2020.

\bibitem{1292216}
Z. {Wang}, E.~P. {Simoncelli}, and A.~C. {Bovik}.
\newblock Multiscale structural similarity for image quality assessment.
\newblock In {\em The Thrity-Seventh Asilomar Conference on Signals, Systems
  Computers, 2003}, volume~2, pages 1398--1402 Vol.2, 2003.

\bibitem{DBLP:conf/cvpr/WebsterRSJ19}
Ryan Webster, Julien Rabin, Lo{\"{\i}}c Simon, and Fr{\'{e}}d{\'{e}}ric Jurie.
\newblock Detecting overfitting of deep generative networks via latent
  recovery.
\newblock In {\em {IEEE} Conference on Computer Vision and Pattern Recognition,
  {CVPR} 2019, Long Beach, CA, USA, June 16-20, 2019}, pages 11273--11282.
  Computer Vision Foundation / {IEEE}, 2019.

\end{thebibliography}
}

\end{document}